\documentclass[11pt]{article}

\usepackage[final]{acl}

\usepackage{times}
\usepackage{latexsym}

\usepackage[T1]{fontenc}

\usepackage[utf8]{inputenc}

\usepackage{microtype}

\usepackage{inconsolata}

\usepackage{graphicx}

\title{\dataset: Benchmarking Complex Mathematical Reasoning of Multimodal Large Language Models Via Error Detection}

\author{\textbf{Yibo Yan}$^{1,2,3}$, 
    \textbf{Shen Wang}$^{1}$, 
    \textbf{Jiahao Huo}$^{2}$, 
    \textbf{Hang Li}$^{1,4}$, 
    \textbf{Boyan Li}$^{2}$, 
    \textbf{Jiamin Su}$^{2}$, 
    \textbf{Xiong Gao}$^{2}$, \\
    \textbf{Yi-Fan Zhang}$^{1,5}$,
    \textbf{Tianlong Xu}$^{1}$,
    \textbf{Zhendong Chu}$^{1}$,
    \textbf{Aoxiao Zhong}$^{1}$,
    \textbf{Kun Wang}$^{1,5}$,\\
    \textbf{Hui Xiong}$^{2,3}$,
    \textbf{Philip S. Yu}$^{6}$,
    \textbf{Xuming Hu}$^{2,3,}$\thanks{Corresponding Authors} ,
    \textbf{Qingsong Wen}$^{1,}$\footnotemark[1]\\
    $^1$Squirrel AI,
    $^2$HKUST(GZ), 
    $^3$HKUST, 
    $^4$MSU,
    $^5$UCAS,
    $^6$University of Illinois at Chicago\\
    \texttt{\href{mailto:yanyibo70@gmail.com}{yanyibo70@gmail.com}},
     \texttt{\href{mailto:xuminghu@hkust-gz.edu.cn}{xuminghu@hkust-gz.edu.cn}},
     \texttt{\href{mailto:qingsongedu@gmail.com}{qingsongedu@gmail.com}},
    \vspace{-3mm}
    \\
    \vspace{-3mm}
}

\usepackage[utf8]{inputenc} 
\usepackage[T1]{fontenc}    
\usepackage{url}            
\usepackage{booktabs}       
\usepackage{amsfonts}       
\usepackage{nicefrac}       
\usepackage{microtype}      
\usepackage{xcolor}
\definecolor{bluecite}{HTML}{0071BC}

\usepackage{graphicx}
\usepackage{subcaption}
\usepackage{multirow}

\usepackage{amsmath}

\usepackage{paralist}

\usepackage[capitalize]{cleveref}
\crefname{section}{Sec.}{Secs.}
\Crefname{section}{Section}{Sections}
\Crefname{table}{Table}{Tables}
\crefname{table}{Tab.}{Tabs.}

\usepackage{microtype}
\usepackage{booktabs} %

\definecolor{darkgreen}{rgb}{0.0, 0.5, 0.0} 

\newcommand{\comt}[1]{#1}

\renewcommand{\comt}[1]{}

\usepackage[dvipsnames,table]{xcolor}

\usepackage{tabularx}
\usepackage{multicol}
\definecolor{myblue}{RGB}{235,235,250}

\usepackage{xspace}

\usepackage{pifont}
\usepackage{latexsym}
\usepackage{inconsolata}
\usepackage{arydshln}
\usepackage{enumitem}
\usepackage{wrapfig}
\usepackage{array}
\usepackage{epigraph}

\definecolor{lightpink}{RGB}{204, 231, 207} 
\definecolor{lightblue}{RGB}{210, 220, 250} 
\definecolor{lightgray}{RGB}{237, 237, 237} 

\definecolor{superlightred}{rgb}{0.99, 0.92, 0.92}
\definecolor{darkgreen}{RGB}{50,100,0}
\definecolor{darkred}{RGB}{200, 0, 0}

\usepackage{makecell}
\usepackage{colortbl}

\usepackage{hyperref}
\usepackage{url}
\usepackage{hyperref}
\usepackage[most]{tcolorbox}
\usepackage{wrapfig}
\usepackage{graphicx,xcolor,float}
\usepackage{subcaption}
\usepackage{threeparttable}
\usepackage{algorithm}
\usepackage{pifont}
\usepackage{colortbl}
\usepackage{color}
\usepackage{multirow}
\usepackage{tabularx}
\usepackage{float}
\usepackage{graphicx}
\usepackage{booktabs}
\usepackage{arydshln}
\usepackage{enumitem}
\usepackage{wrapfig}
\usepackage{caption}
\usepackage{graphicx}
\usepackage{makecell}
\usepackage{float} 
\usepackage{makecell}
\usepackage{tabularx}
\usepackage{twemojis}
\usepackage{amssymb,mathrsfs,amsmath}
\usepackage{pifont}
\usepackage[export]{adjustbox}
\usepackage{xcolor}
\usepackage{xspace}
\usepackage{shadowtext}
\usepackage{anyfontsize}

\hypersetup{
    colorlinks=true,
    linkcolor=red,
    citecolor=cyan,
    filecolor=magenta,      
    urlcolor=magenta,
    }

\definecolor{bittersweet}{rgb}{1.0, 0.44, 0.37}
\definecolor{mygreen}{rgb}{0.29, 0.7, 0.48}
\definecolor{my_green}{RGB}{51,102,0}
\definecolor{my_yellow}{RGB}{255,165,0}
\definecolor{my_red}{RGB}{204, 0, 0}
\definecolor{my_orange}{RGB}{232, 132, 23}

\newcommand{\red}[1]{\textcolor{red}{#1}}

\newcommand{\green}[1]{\textcolor{my_green}{#1}}

\newcommand{\blue}[1]{\textcolor{blue}{#1}}
\newcommand{\orange}[1]{\textcolor{my_orange}{#1}}
\usepackage{pifont}
\definecolor{mygray}{gray}{0.4}
\hypersetup{
    colorlinks=true,
    linkcolor=red,
    citecolor=cyan,
    filecolor=magenta,      
    urlcolor=magenta,
    }
\usepackage{xcolor} 
\usepackage{xcolor}

\usepackage[font=small,labelfont=bf]{caption}  
\usepackage{makecell}
\usepackage{tabulary}
\definecolor{ada_green}{rgb}{0,205,205}
\definecolor{glt_red}{rgb}{109,205,255}

\newcommand{\header}[1]{\text{#1}}

\definecolor{backred}{RGB}{255, 190, 190}
\definecolor{backblue}{RGB}{210, 230, 250}
\definecolor{backgrey}{RGB}{220, 220, 220}
\newcommand{\high}{\cellcolor{backblue}}

\newcommand{\best}{\cellcolor{backred}}

\definecolor{shadecolor}{RGB}{237,237,237}

\newcommand{\dataset}{\textsc{ErrorRadar}\xspace}

\usepackage[toc]{appendix}
\usepackage{etoc}
\usepackage{minitoc}
\etocsettocstyle{\section*{Contents of Technical Appendices}}{}

\begin{document}
\maketitle

\etocdepthtag.toc{mtchapter}
\etocsettagdepth{mtchapter}{subsection}
\etocsettagdepth{mtappendix}{none}

\begin{abstract}
As the field of Multimodal Large Language Models (MLLMs) continues to evolve, their potential to handle mathematical reasoning tasks is promising, as they can handle multimodal questions via cross-modal understanding capabilities compared to text-only LLMs. Current mathematical benchmarks \textit{predominantly focus on evaluating MLLMs' problem-solving ability}, yet there is a crucial gap in addressing more complex scenarios such as error detection, for enhancing reasoning capability in complicated settings. To fill this gap, we formally formulate the new task — \textbf{multimodal error detection}, and introduce \textbf{\dataset}, the first benchmark designed to assess MLLMs' capabilities in such a task. \dataset evaluates two sub-tasks: \textit{error step identification} and \textit{error categorization}, providing a framework for evaluating MLLMs' complex mathematical reasoning ability. It consists of 2,500 high-quality multimodal K-12 mathematical problems, collected from real-world student interactions in an educational organization, with expert-based annotation and metadata such as problem type and error category. Through extensive experiments, we evaluated both open-source and closed-source representative MLLMs, benchmarking their performance against educational expert evaluators. Results indicate challenges still remain, as GPT-4o with the best model performance is around 10\% behind human evaluation. The dataset can be found in this \href{https://huggingface.co/datasets/ErrorRadar/ErrorRadar}{repo}, with version iteration in the future.
\end{abstract}

\section{Introduction}
\label{sec:introduction}

On the path to Artificial General Intelligence, Large Language Models (LLMs) such as GPT-4 \citep{openai2023gpt4} have emerged as a central focus in both industry and academia \citep{minaee2024large, zhao2023survey, zhu2023large,yan2026unlocking}. As the real world is inherently multimodal, the evolution of Multimodal Large Language Models (MLLMs) such as the latest GPT-4o \citep{openai2024gpt4o} and Gemini series \citep{reid2024gemini}, has become a growing area of interest, demonstrating remarkable effectiveness in diverse applications \citep{xiao2024comprehensive, he2024pefomed, yan2024urbanclip, hao2024urbanvlp}. In particular, multimodal reasoning stands to benefit education scenarios from the robust capabilities of MLLMs \citep{wang2024large, li2024bringing}, given its reliance on multimodal inputs to comprehensively grasp users' intentions.

\begin{figure}[t!]
    \centering
    \includegraphics[width=0.9 \linewidth]{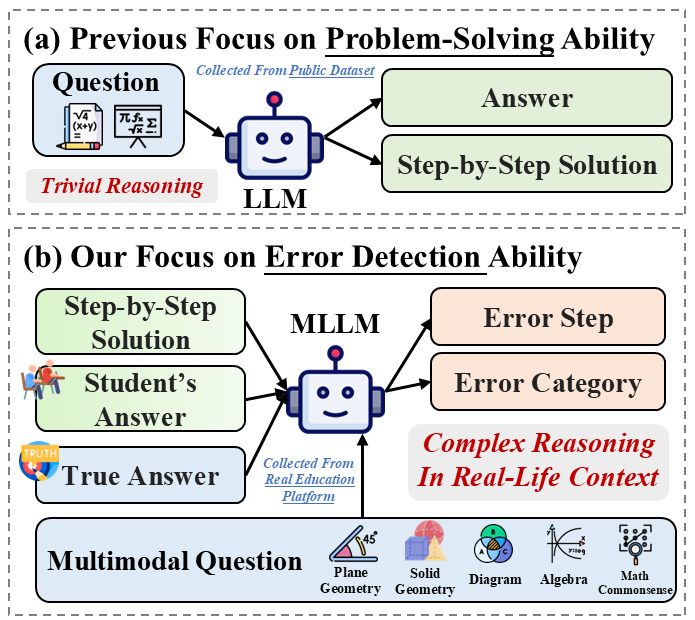}
    \caption{Comparison of research scope and task setting between previous works (a) and our proposed \dataset benchmark (b) on mathematical reasoning tasks.}
    \label{fig:scope comparison}
    \vspace{-6mm}
\end{figure}

Within the multimodal sphere, mathematical scenarios pose a significant challenge, demanding sophisticated reasoning abilities from MLLMs \citep{ahn2024large,yan2024survey,huang2025effireason}. These scenarios have attracted considerable research aimed at pushing the boundaries of MLLMs' reasoning capabilities \citep{hu2024visual,jia2024describe,lu2024chameleon,shi2024math,zhuang2024math}. Besides, various representative benchmarks have been designed to measure MLLMs' performance in complex mathematical reasoning tasks, which involve multi-step reasoning and quantitative analysis within visual contexts \citep{lu2023mathvista,peng2024multimath}.

Scrutinizing the off-the-shelf mathematical reasoning benchmarks, there is a predominant focus on evaluating the problem-solving capabilities of MLLMs, prioritizing the accuracy with which MLLMs can solve mathematical problems \citep{wang2024measuring,lu2023mathvista}, as depicted in Figure \ref{fig:scope comparison} (a). However, in educational contexts, it is even more crucial to consider user-oriented needs, such as \textbf{error detection}. As indicated in Figure \ref{fig:scope comparison} (b), this involves not only pinpointing the first incorrect step in a student's step-by-step solution but also categorizing the types of errors made, which is a multifaceted process that requires a deep understanding of both mathematical concepts and cognitive processes \citep{rabillas2023elementary}.

\begin{figure}[t!]
    \centering
        \begin{table}[H]
            \centering
            \begin{adjustbox}{width=0.45\textwidth}
            \begin{tabular}{ccccccc}
                \toprule
                \textbf{Benchmarks} & \textbf{Venue} & \textbf{Modality} & \textbf{Student Ans.} & \textbf{Error Det.} \\
                \midrule
                 \multicolumn{1}{l}{TheoremQA \citep{chen2023theoremqa}}
                & EMNLP & $T$ & - & -  \\ 
                \multicolumn{1}{l}{MathBench \citep{liu2024mathbench}}
                & ACL & $T$ & - & -  \\ 
                \multicolumn{1}{l}{MR-GSM8K \citep{zeng2024mr}}
                & ICLR & $T$  & - & -  \\ 
                \multicolumn{1}{l}{SciEval \citep{sun2024scieval}}
                & AAAI & $T$  & - & - \\ 
                \multicolumn{1}{l}{EIC \citep{li2024evaluating}}
                & ACL Finding & $T$  & - & \checkmark   \\ 
                \multicolumn{1}{l}{CMMaTH \citep{li2024cmmath}}
                & COLING & $T,I$ & - & -  \\ 
                \multicolumn{1}{l}{MathScape \citep{zhou2024mathscape}}
                & arXiv & $T,I$ & - & -  \\ 
                \multicolumn{1}{l}{MATH-V \citep{wang2024measuring}}
                & NeurIPS & $T,I$ & - & -  \\ 
                \multicolumn{1}{l}{QRData \citep{liu2024llms}}
                & ACL & $T,I$ & - & -  \\ 
                \multicolumn{1}{l}{IsoBench \citep{fu2024isobench}}
                & COLM & $T,I$ & - & -  \\ 
                \multicolumn{1}{l}{SciBench \citep{wang2023scibench}}
                & ICML & $T,I$ & - & -  \\ 
                \multicolumn{1}{l}{MathVista \citep{lu2023mathvista}}
                & ICLR & $T,I$ & - & -   \\ 
                \multicolumn{1}{l}{MathVerse \citep{zhang2024mathverse}}
                & ECCV & $T,I$  & - & - \\ 
                \rowcolor{gray!40} \multicolumn{1}{l}{\dataset (Ours)}
                & - & $T,I$  & \checkmark & \checkmark \\ 
                \bottomrule
            \end{tabular}
            \end{adjustbox}
            \caption{Comparison between our proposed \dataset vs. representative LLM-based mathematical reasoning benchmarks. $T$ and $I$ represent text and image. \textbf{\textit{Student Ans.}} indicates if the dataset contains real student data (\textit{i.e.}, students' incorrect answers); \textbf{\textit{Error Det.}} represents if error detection task is included. See more comparison in Appendix \ref{app:bench_comprison}.}
            \label{tab:bench comparison}
            \vspace{-2.5em}
        \end{table}    
\end{figure}

Towards this end, addressing the aforementioned research gap, we aim to formulate the new task of evaluating MLLMs in the context of error detection scenarios, and therefore introduce the corresponding benchmark termed \textbf{\dataset}. We have designed two sub-tasks to comprehensively assess the performance: \textit{error step identification} and \textit{error categorization}. To construct a rich and reliable dataset, we initially sourced a collection of multimodal K-12 level math problems from an educational organization and subsequently refined the dataset through rigorous manual annotation to ensure quality. In particular, we also collect real students' answers for each multimodal question for a relatively robust experimental setting, compared to other relevant benchmarks (See comparisons in Figure \ref{tab:bench comparison}). Furthermore, we categorized the dataset to better align with diverse needs as follows: \textbf{Problem types}: \textit{plane geometry}, \textit{solid geometry}, \textit{diagram}, \textit{algebra}, and \textit{mathematical common sense}; and \textbf{Error categories}: \textit{visual perception errors}, \textit{calculation errors}, \textit{reasoning errors}, \textit{knowledge errors}, and \textit{misinterpretation of the problem}. In summary, \dataset comprises 2,500 high-quality instances derived from real-life problem-solving data, providing a foundational dataset to enhance the complex reasoning capabilities of MLLMs for the research community and industry.

For \dataset, we carry out an extensive experimental analysis to determine the proficiency in complex mathematical reasoning of various MLLMs. The evaluation encompasses both the latest open-source MLLMs (\textit{e.g.}, InternVL2 \citep{chen2023internvl}, LLaVA-NEXT \citep{liu2024llavanext}, CogVLM2 \citep{wang2023cogvlm}), and closed-source MLLMs (\textit{e.g.}, GPT4-o \citep{openai2024gpt4o}, Gemini Pro 1.5 \citep{reid2024gemini}, Claude 3.5 \citep{claude35}). Our focus was on \textit{their error detection capabilities, specifically the identification of the erroneous step and the classification of the error type.} To establish a comparative human performance standard, we involved expert human educators who possess a graduate-level degree or higher qualifications.
The results demonstrate that \dataset, covering cutting-edge topics such as MLLMs' complex reasoning, poses a significant challenge, with human evaluation for two error detection tasks achieving less than 70\%.

From in-depth evaluation of representative MLLMs, we obtain the following findings: \ding{182} Closed-source MLLMs, particularly GPT-4o, consistently outperform open-source MLLMs in both sub-tasks, and show more balanced performance across different error categories; \ding{183} Weaker MLLMs exhibit an over-reliance on simpler categories, while stronger models handle complex tasks better; \ding{184} Both MLLMs and humans perform better on error step identification compared to error categorization, as localizing specific errors is inherently simpler than categorizing errors. Our contributions can be summarized as follows\footnote{Refer to Appendix \ref{app:impact_statement} and \ref{app:task_clarification} for the impact statement and more clarification of our proposed new task setting.}: 

\begin{itemize}[leftmargin=*]
    \item[\ding{182}] We take the \textbf{first step to formulate the multimodal error detection task}, and introduce a multimodal benchmark termed \dataset. This benchmark serves as a standard operator for assessing the complex mathematical reasoning capabilities of the latest MLLMs.
    
    \item[\ding{183}] We meticulously curate an extensive dataset comprising approximately 2,500 high-quality instances with rigorous annotation and rich metadata derived from real user interactions in an educational organization. To the best of our knowledge, this is the first attempt to use real-world student problem-solving data to evaluate MLLMs.
    
    \item[\ding{184}] Our comprehensive experimental evaluation of more than 20 MLLMs, both proprietary and open-source, highlight the substantial room for improvement (\textit{i.e.}, 7\%-15\% in performance) in the complex mathematical reasoning capabilities, underscoring the necessity for further research.
\end{itemize}

\section{Related Work}
\label{sec:related_work}

\textbf{Benchmarks for Mathematical Reasoning.} 
Recent advancements in mathematical reasoning benchmarks have led to the development of both pure text and multimodal assessments \citep{lu2022survey,wang2024measuring,zheng2024reefknot,huo2024mmneuron}. While datasets like GSM8K \citep{cobbe2021training}, MATH \citep{hendrycks2021measuring}, SuperCLUE-Math \citep{xu2024superclue}, and MathBench \citep{liu2024mathbench} focus on text-based problems, the field has expanded to include multimodal benchmarks that introduce visual elements, pushing the boundaries of AI's mathematical understanding. For instance, MathVista \citep{lu2023mathvista} evaluates AI's performance on visual math QA tasks; MATH-V \citep{wang2024measuring} focuses on multimodal mathematical understanding with competition-derived questions; MathVerse \citep{zhang2024mathverse} assesses visual diagram comprehension using CoT strategies; CMMU \citep{he2024cmmu} tests multi-disciplinary, multimodal math understanding with a broad range of Chinese-language questions; MathScape \citep{zhou2024mathscape} further advances the field by presenting longer, more complex, and open-ended multimodal problems; and MMMU \citep{yue2024mmmu} covers college-level knowledge including interleaved mathematical questions. The aforementioned benchmarks assess the mathematical reasoning capabilities of MLLMs by evaluating their problem-solving levels, but they overlook tasks based on the student's perspective, such as error detection. Therefore, we propose the \dataset benchmark, entirely based on real student response data. We discuss more relevant reasoning benchmarks and specific MLLMs in Appendix \ref{app:more_related_work}.

\textbf{Multimodal Large Language Models.} 
Generative foundation models such as GPT-4~\citep{openai2023gpt4}, Claude~\citep{claude35}, and Gemini~\citep{pal2024gemini} have significantly advanced various task solutions without fine-tuning \citep{cui2024survey,yan2024georeasoner,zou2025deep,zhong2024urbancross}. 
Similarly, current open-source MLLMs, built on top of powerful LLMs, have also demonstrated promising potential in multimodal tasks such as image captioning~\citep{yang2024exploring} and visual question answering~\citep{fan2024muffin}. For instance, LLaVA-NEXT~\citep{liu2024llavanext} proposed projecting visual embeddings, extracted by a pretrained vision encoder, into the word space through a single MLP layer, where LLMs like LLaMA, Vicuna, and Mistral are fine-tuned to understand these post-projection tokens. In a similar fashion, Phi3~\citep{abdin2024phi3}, DeepSeek-VL~\citep{lu2024deepseek}, MiniCPM-V~\citep{yao2024minicpm}, ChatGLM~\citep{glm2024chatglm}, CogVLM~\citep{wang2023cogvlm}, Intern-VL~\citep{chen2023internvl}, Qwen-VL~\citep{Qwen-VL} and Yi-VL~\citep{young2024yi} also utilize a projector (or adapter, shared compression layer, \textit{etc.}) to align the visual embeddings extracted from a vision encoder with text embeddings, which are then concatenated and fed into LLM. 
Therefore, we propose \dataset, a benchmark on a fine-grained evaluation of MLLMs' ability to detect errors based on students' answers and reasoning steps.

\section{The \dataset Dataset}
\label{sec:dataset}

\subsection{Task Formulation}
\label{sec:task_definition}

\begin{figure*}[t!]
  \centering
  \includegraphics[width=0.8\textwidth]{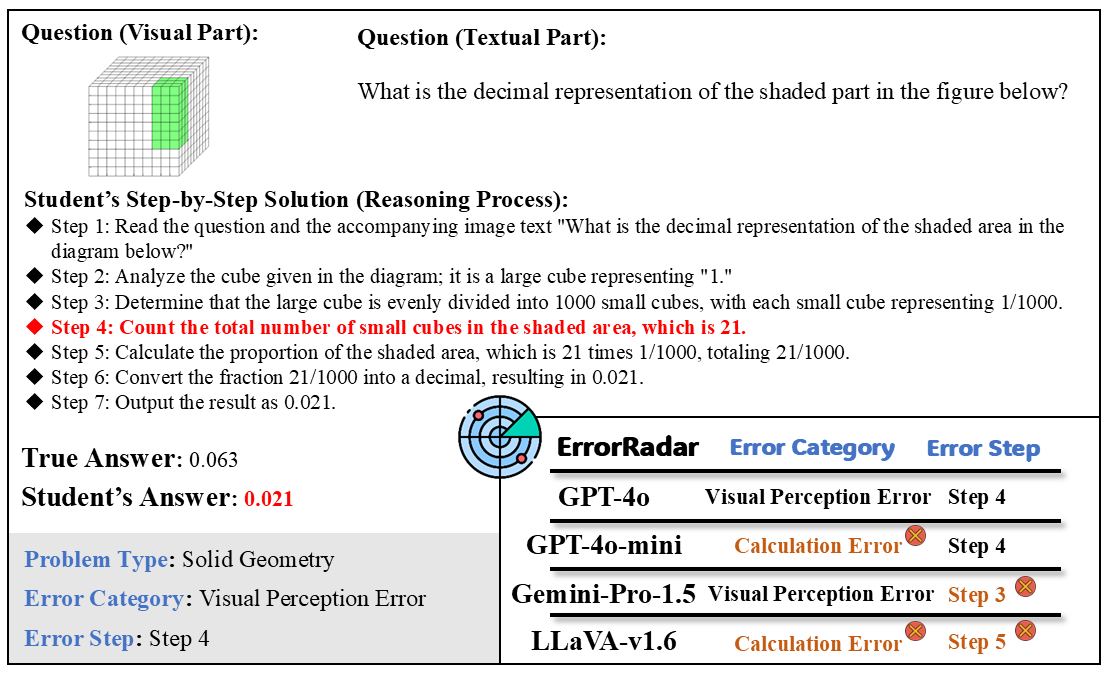}
  \caption{Example of our annotated multimodal mathematical reasoning dataset \dataset, and performance comparison on error categorization and error step localization tasks among representative MLLMs. It is evident even simple math problems can be mishandled by superior MLLMs in one or both tasks, highlighting the challenging nature of multimodal error detection.}
  \vspace{-4mm}
\label{fig:task definition}
\end{figure*}
\textbf{Basic Setting.} 
In this task, we assess the model's ability to detect errors in mathematical problem-solving processes across multiple samples. Let $N$ denote the total number of samples in the evaluation set. For each sample $i \in \{1, 2, \dots, N\}$, the input set $\mathcal{I}_i$ is defined as:
\[
\mathcal{I}_i = \{Q_{\text{text}, i}, Q_{\text{image}, i}, A_{\text{correct}, i}, A_{\text{incorrect}, i}, \{S_{k, i}\}_{k=1}^{n_i}\},
\]
where $Q_{\text{text}, i}$ represents the textual statement of the $i$-th problem, $Q_{\text{image}, i}$ represents the image representation of the $i$-th problem, $A_{\text{correct}, i}$ represents the correct solution for the $i$-th problem, $A_{\text{incorrect}, i}$ represents the incorrect student solution for the $i$-th problem, and $\{S_{k, i}\}_{k=1}^{n_i}$ denotes the sequence of $n_i$ steps in the $i$-th problem-solving process, with each $S_{k, i}$ representing a distinct step.

\textbf{Subtask 1: Error Step Identification.}
The task is to identify the index $x$ of the first incorrect step in the sequence $\{S_{k, i}\}_{k=1}^{n_i}$. The function $f_{\text{step}, i}$ maps the input $\mathcal{I}_i$ to the index of the erroneous step:
\[
f_{\text{step}, i} : \mathcal{I}_i \rightarrow x_i, 
\]
\[
\quad \text{where } x_i = \arg\min_{k} \{S_{k, i} \ \text{is incorrect}\}.
\]
\vspace{-0.1em}
\textbf{Subtask 2: Error Categorization.}
The task is to classify the type of error for the $i$-th problem into one of the following categories:
$\{\text{VIS}, \text{CAL}, \text{REAS}, \text{KNOW}, \text{MIS}\}$. The error categorization function $f_{\text{error}, i}$ maps the input $\mathcal{I}_i$ to the error category $C_{\text{error}, i}$:
\[
f_{\text{error}, i} : \mathcal{I}_i \rightarrow C_{\text{error}, i}.
\]
More concrete examples can be seen in Figure \ref{fig:task definition} and Appendix \ref{app:example}. The discrepancies within the five error categories are delineated as follows:
\begin{itemize}[leftmargin=*]
    \item[\ding{79}]  \textbf{Visual Perception Errors (VIS)}: These errors arise when there is a failure to accurately interpret the information contained in images or diagrams presented in the question due to visual issues.
    \item[\ding{79}]  \textbf{Calculation Error (CAL)}: These errors manifest during the calculation process, which may include arithmetic mistakes such as incorrect addition, subtraction, multiplication, or division, errors in unit conversion, or mistakes in the numerical signs between multiple steps.
    \item[\ding{79}]  \textbf{Reasoning Error (REAS)}: These errors occur during problem-solving process when improper reasoning is applied, leading to incorrect application of logical relationships or conclusions.
   \item[\ding{79}]  \textbf{Knowledge Error (KNOW)}: These errors result from incomplete or incorrect understanding of the knowledge base, leading to mistakes when applying relevant knowledge points.
    \item[\ding{79}]  \textbf{Misinterpretation of the Question (MIS)}: These errors occur when there is a failure to correctly understand the requirements of the question or a misinterpretation of the question's intent, leading to responses that are irrelevant to the question's demands. 
\end{itemize}
\textbf{Performance Metrics.}
The evaluation of both subtasks is conducted separately:
\vspace{-0.8em}
\begin{itemize}[leftmargin=*]
    \item \textbf{Error Step Identification Performance.} Let $G_{\text{step}, i}$ be the ground truth index of the first incorrect step for the $i$-th sample. We report the accuracy for this subtask:
    \[
    \text{Acc}_{\text{step}} = \frac{1}{N} \sum_{i=1}^{N} \mathbb{I}(x_i = G_{\text{step}, i}),
    \]
    where $\mathbb{I}(\cdot)$ is indicator function, returning 1 if prediction matches ground truth, and 0 otherwise.
    \item \textbf{Error Categorization Performance.} We report Precision, Recall, and F1-score for each error category, alongside their macro-averaged counterparts as overall performance metrics.
\end{itemize}
\vspace{-0.8em}
\subsection{Data Source \& Annotation}
\label{sec:data_source}

\begin{figure}[!t]
    \centering
    \includegraphics[width=1\linewidth]{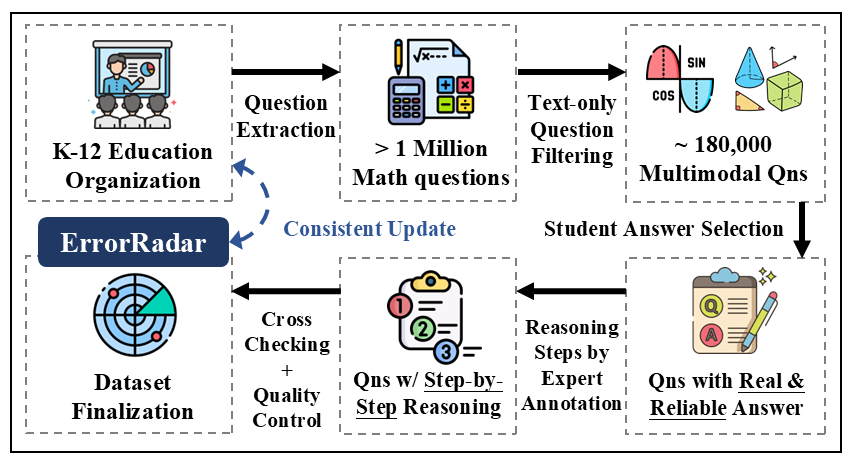}
    \caption{Roadmap of \dataset dataset collection, filtering, annotation, and consistent update.}
    \label{fig:dataset_roadmap}
    \vspace{-2mm}
\end{figure}

\vspace{-2mm}
Following the roadmap shown in Figure \ref{fig:dataset_roadmap}, this section includes how we collect and annotate \dataset dataset to ensure the overall data quality. Different from the conventional benchmarks that rely on public datasets or modified textbook collections \citep{lu2023mathvista,zhou2024mathscape}, \dataset dataset is uniquely sourced from the question bank of a global educational organization. This repository encompasses a vast array of mathematical problems in K-12 levels, totaling over a million entries. Initially, we curated approximately 180,000 math problems with a single-image setting, aligning with multimodal setup.

Subsequently, we refined our selection by evaluating the universality and articulation of the problem content. For each problem, we identified multiple incorrect answers, and finally selected the most frequently given incorrect answer as the student's response. Additionally, we scrutinized cases where the most common incorrect answer was due to system input errors despite the answer being correct. In such instances, we amended the dataset by incorporating the next most frequently incorrect answer. 

\begin{figure}[!t]
    \centering
    \includegraphics[width=1\linewidth]{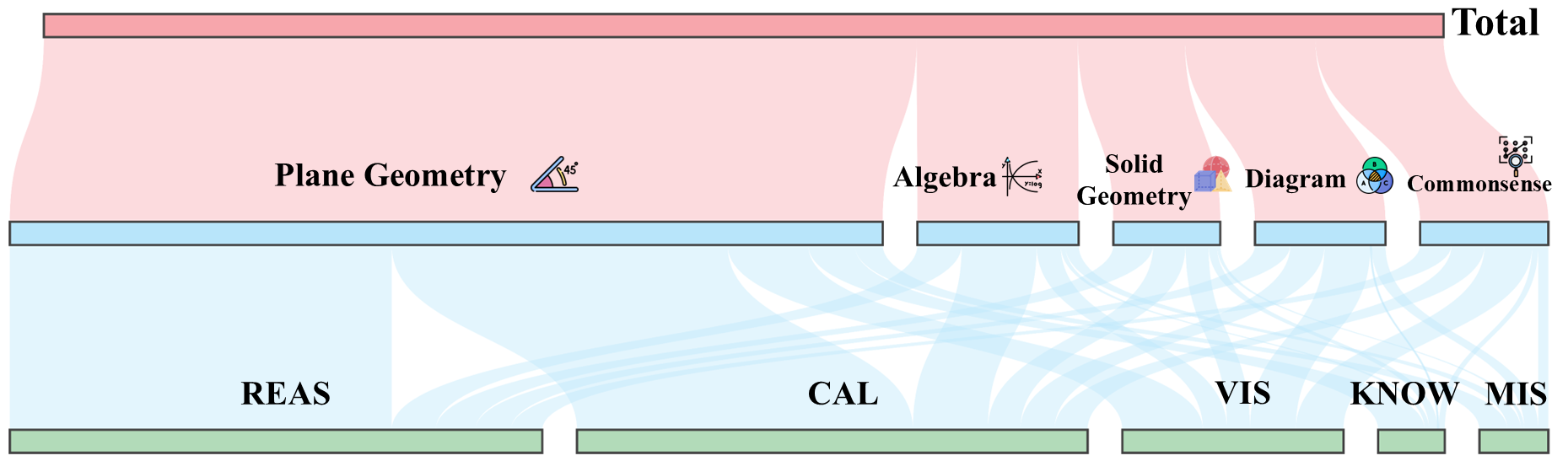}
    \caption{Dataset distribution of \dataset with respect to problem type and error category.}
    \label{fig:dataset_flow}
    \vspace{-4mm}
\end{figure}

Furthermore, since error detection tasks necessitate a step-by-step reasoning process, we enriched our dataset with new content through manual annotation. Specifically, we provided professional annotators with the original multimodal QA data, student's incorrect answers, and the pedagogical team's analysis of correct answer process. Based on initial data, annotators delineated the erroneous steps leading to the incorrect answers (More details of annotation and inconsistent case handling in Appendix \ref{app:annotation detail} and \ref{app:annotation inconsistency handling}).
\vspace{-0.7mm}

\begin{figure}[!t]
    \centering
    \fontsize{7.2pt}{\baselineskip}\selectfont 
    \renewcommand\tabcolsep{0.8pt} 
    \renewcommand\arraystretch{0.6} 
    \begin{tabular}[t]{lc} 
        \toprule
        \textbf{Statistic} & \textbf{Number} \\
        \midrule
        Total multimodal questions & 2,500 \\
        \midrule
        Problem Type &  \\
        ~- Plane Geometry & 1559 (62.4\%) \\
        ~- Solid Geometry & 191 (7.6\%) \\
        ~- Diagram & 233 (9.3\%) \\
        ~- Algebra & 288 (11.5\%) \\
        ~- Math Commonsense & 229 (9.2\%) \\
        \midrule
        Error Category &  \\
        ~- Visual Perception Error & 395 (15.8\%) \\
        ~- Calculation Error & 912 (36.5\%) \\
        ~- Reasoning Error & 951 (38.0\%) \\
        ~- Knowledge Error & 119 (4.8\%) \\
        ~- Misinterpretation of the Qns & 123 (4.9\%) \\
        \midrule
        Average Reasoning Step  & 7.6 \\
        Maximum Reasoning Step & 20 \\
        Minimum Reasoning Step & 3 \\
        Average Question Length  & 168 \\
        Maximum Question Length & 719 \\
        Minimum Question Length & 13 \\
        \bottomrule
    \end{tabular}
    \caption{Key statistics of \dataset.}
    \label{tab:statistics}
    \vspace{-4mm}
\end{figure}

Our team of annotators, consisting of around ten educational experts with domain expertise, conducted two rounds of cross-checking to ensure the reliability of the annotations. In cases of inconsistency, the related data were presented to the annotation lead for final adjudication. The annotators' results were subject to review and quality control by the educational organization from which the data originated, ensuring \textit{security}, \textit{reliability}, and \textit{consistent updates}.
\vspace{-2mm}
\subsection{Dataset Details}
\label{sec:dataset_detail}
As illustrated in Figure \ref{tab:statistics}, \dataset dataset comprises a substantial collection of 2,500 multimodal math questions designed for error detection tasks. It predominantly includes plane geometry problems, with solid geometry, diagram, algebra, and math commonsense questions making up the remainder, highlighting its focus on diverse mathematical problems. It also categorizes errors into visual perception, calculation, reasoning, knowledge, and question misinterpretation. Key statistics indicate a diverse dataset with an average reasoning step of 7.6, a variety of question lengths, and a wide range of reasoning steps. Detailed distribution of \dataset, problem type definition, and error category formulation can be seen in Figure \ref{fig:dataset_flow}, Appendix \ref{app:problem type cate} and \ref{app:error category finalization process}.
\section{Experiments and Analysis}
\label{sec:experiment}

\subsection{Evaluation Protocols}
\label{sec:eval_protocols}
In \dataset benchmark, we propose an evaluation strategy using template matching rules. The evaluation process consists of three stages: \textit{response generation}, \textit{answer extraction}, and \textit{performance calculation}. Initially, the MLLMs generate responses given the inputs, which incorporates the multimodal mathematical question, wrong answer, and its step-by-step reasoning, using the template from Appendix \ref{app:prompt}. Subsequently, the short answer text can be extracted from the detailed response. Finally, the model performance is based on the detailed score calculation as shown in Section \ref{sec:task_definition}. The final score will be calculated by averaging the scores from three rounds of assessment.
\vspace{-2mm}

\subsection{Experimental Setup}
\label{sec:experiment_setup}

\begin{figure*}[t!]
  \centering
  \includegraphics[width=0.8\textwidth]{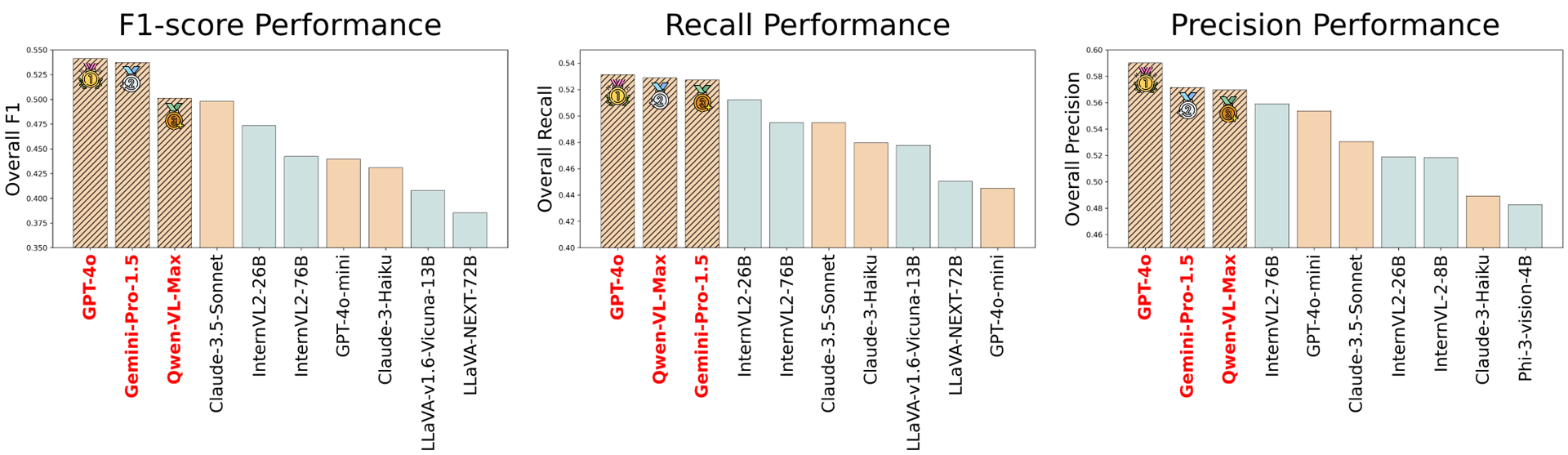}
  \vspace{-3mm}
  \caption{Error category performance of top 10 MLLMs for F1, recall, and precision, respectively. The \orange{orange} bars represent closed-source MLLMs, while the \blue{blue} ones represent open-source MLLMs. The masked bars represent the top 3. Due to page limit, we leave the bar charts of all models' performance in Appendix \ref{app:main_result}.}
  \vspace{-6mm}
\label{fig:cate_performance_bar_top10}
\end{figure*}

In our experimental setup, we meticulously categorized and evaluated a diverse array of MLLMs into three distinct groups to assess their capabilities across error detection tasks. (i) The \textbf{Open-Source MLLMs} category encompassed models such as InternVL-2 \citep{chen2023internvl}, Phi-3-vision \citep{abdin2024phi3}, Yi-VL \citep{young2024yi}, DeepSeek-VL \citep{lu2024deepseek}, LLaVA-v1.6-Vicuna \citep{liu2024llavanext}, MiniCPM-LLaMA3-V2.5 \citep{yao2024minicpm}, MiniCPM-V2.6 \citep{yao2024minicpm}, Qwen-VL \citep{Qwen-VL}, GLM-4v \citep{glm2024chatglm}, and LLaVA-NEXT \citep{liu2024llavanext}, each demonstrating their unique strengths and capabilities in handling different types of errors. (ii) The \textbf{Closed-Source MLLMs} featured proprietary models like Qwen-VL-Max \citep{Qwen-VL}, Claude-3-Haiku \citep{claude3}, Claude-3.5-Sonnet \citep{claude35}, Gemini-Pro-1.5 \citep{reid2024gemini}, GPT-4o-mini \citep{openai2024gpt4omini}, and GPT-4o \citep{openai2024gpt4o}, providing a comparison point for the performance of models that are not publicly accessible. (iii) Lastly, the \textbf{Human Performance} category served as a benchmark for natural intelligence, allowing us to gauge how closely MLLMs can emulate human cognitive functions across tasks such as visual perception (More details in Appendix \ref{app:human}). Prompts for MLLMs and sources of MLLMs are in Appendix \ref{app:prompt} and \ref{app:sources}, respectively.

\subsection{Experimental Results}
\label{sec:experiment_result}

\subsubsection{Main Results}
\label{sec:main_result}

\begin{figure}[!t]
    \centering
    \includegraphics[width=1\linewidth]{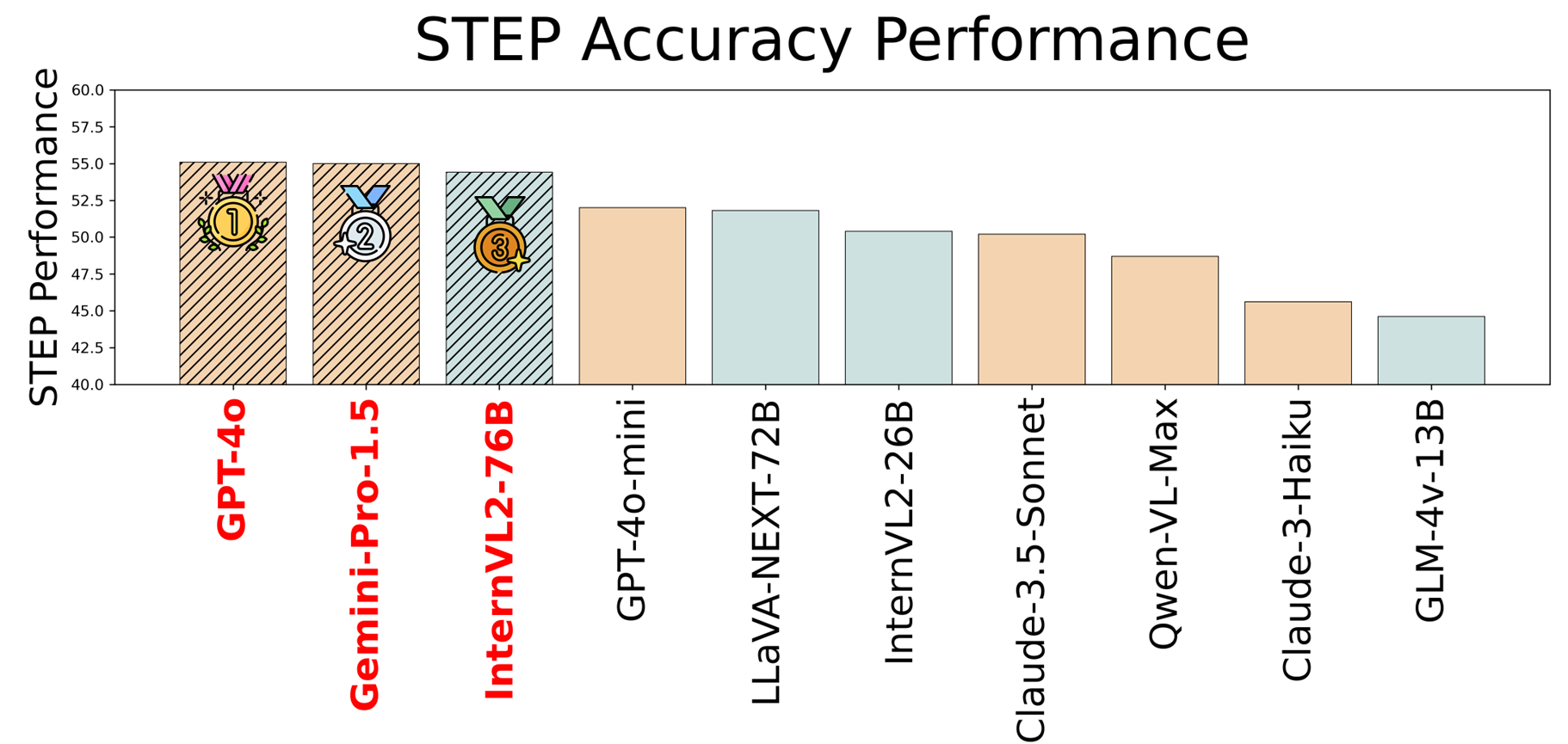}
    \caption{Error step performance of top 10 MLLMs. The \orange{orange} bars represent closed-source MLLMs, while the \blue{blue} ones represent open-source MLLMs. The masked bars represent the top 3. We leave the bar charts of all models' performance in Appendix \ref{app:main_result}.}
    \label{fig:step_performance_bar_top10}
    \vspace{-4mm}
\end{figure}

\begin{figure*}[t!]
  \centering
  \includegraphics[width=0.8\textwidth]{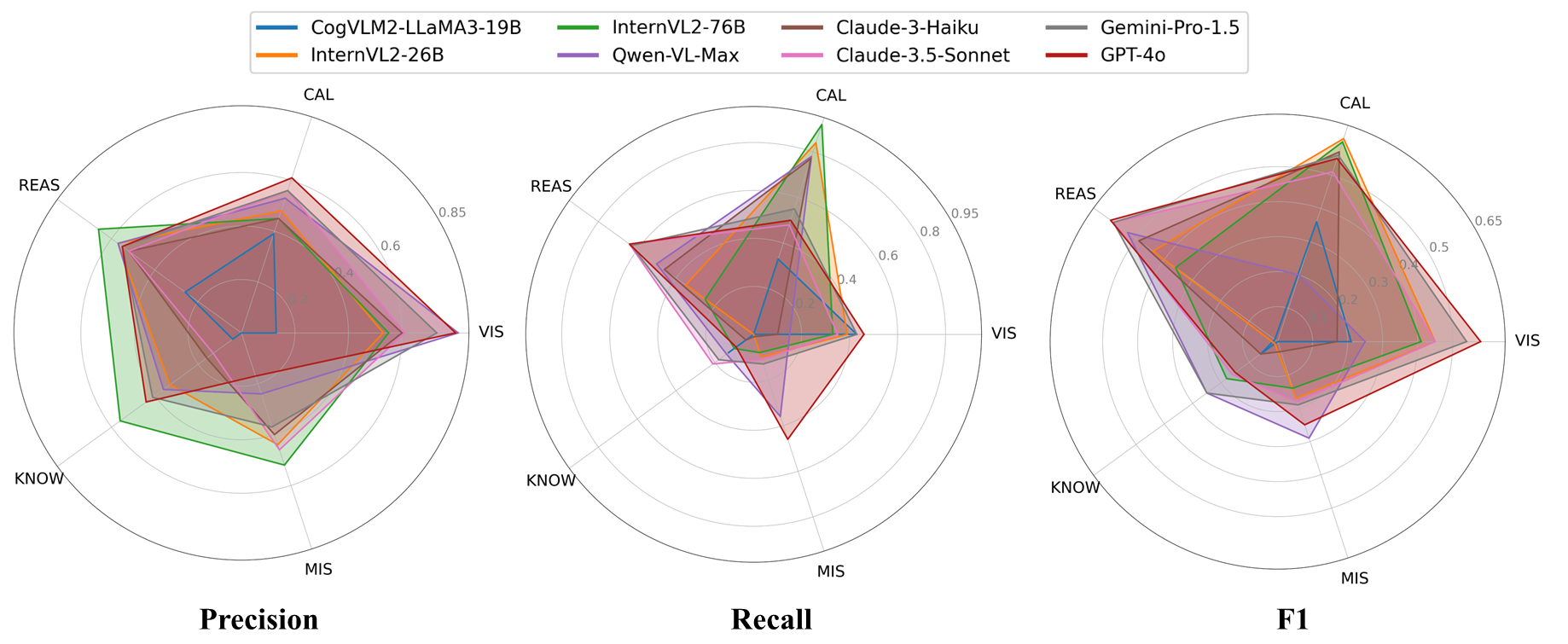}
  \vspace{-3mm}
  \caption{The radar charts of error category performance for the top eight MLLMs (each dimension indicates an error category). Considering visualization clarity, we leave the detailed values of all models in Appendix \ref{app:main_result}.}
\label{fig:main_cate_category_result}
\end{figure*}

\textbf{Finding \#1: Closed-source MLLMs generally outperform open-source MLLMs in both error detection tasks, with GPT-4o demonstrating the strongest performance.} Figures \ref{fig:cate_performance_bar_top10} and \ref{fig:step_performance_bar_top10} show that closed-source MLLMs generally outperform open-source MLLMs in both STEP and CATE tasks, and they also exhibit relatively more balanced performance across the five error categories. This superiority can likely be attributed to the proprietary datasets and advanced training resources available to closed-source models, which allow for more robust fine-tuning \citep{shi2023detecting,yu2024large}. Notably, GPT-4o stands out as the best model, achieving highest scores not only in STEP and CATE tasks, demonstrating its overall versatility. Given the performance gap, open-source MLLMs can further enhance themselves by distilling error detection capabilities of closed-source ones \citep{hsieh2023distilling}. See more actionable suggestions in Appendix \ref{app:actionable suggestions}.


\begin{figure}[!t]
    \centering
    \includegraphics[width=0.8\linewidth]{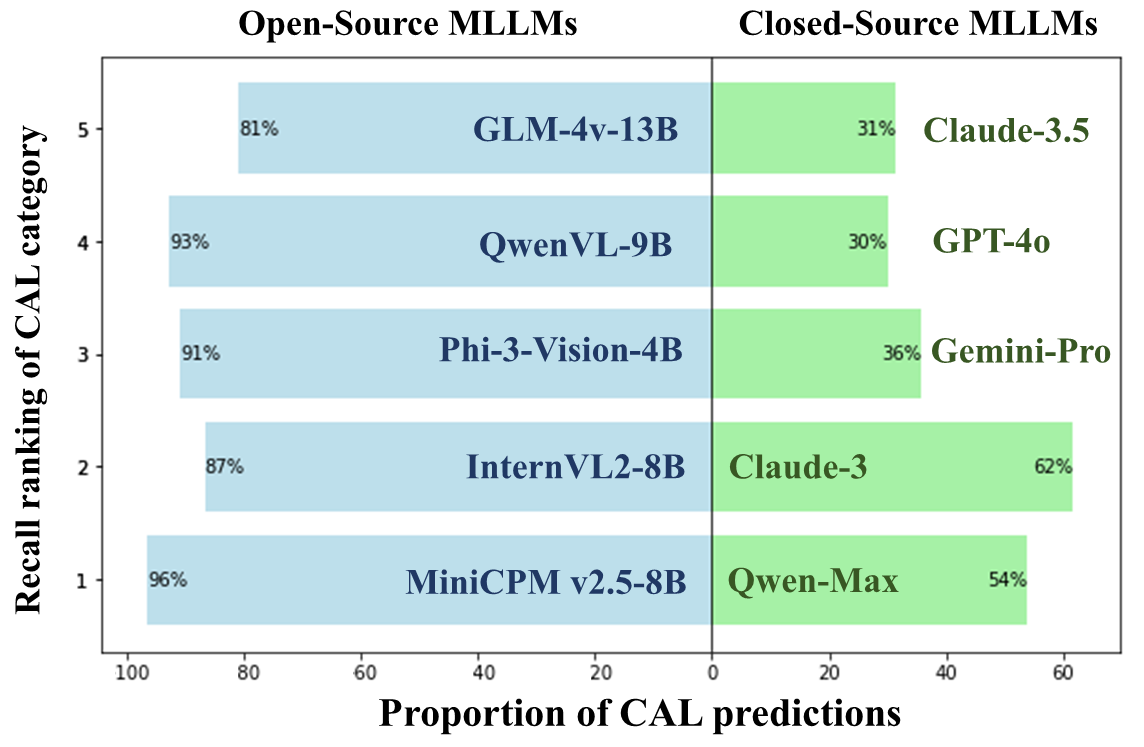}
    \caption{The proportion of CAL predictions of respective representative \green{closed-source} and \blue{open-source} MLLMs.}
    \label{fig:cal_bias}
    \vspace{-2mm}
\end{figure}

\textbf{Finding \#2: Open-source MLLMs tend to predict CAL category, leading to unusually high recall.}\label{obs:harder} Figure \ref{fig:main_cate_category_result} indicates that MLLMs with relatively low performance in the CATE task tend to exhibit unusually high recall in the CAL category. Specifically, open-source models like MiniCPM-LLaMA3-v2.5 even achieve a 100\% recall in CAL, while Phi-3-vision and InternVL-2-8B reach 99.6\%. Upon analyzing the category prediction proportions of CAL from Figure \ref{fig:cal_bias} (See details of all MLLMs in Appendix \ref{app:cal_distribution}), it becomes clear that open-source MLLMs with the top five CAL recall predict over 80\% of instances as  CAL category, suggesting an over-reliance on this category. In contrast, closed-source MLLMs with top-five CAL recall do not exhibit this extreme trend of prediction bias. This phenomenon likely arises from weaker MLLMs attempting to overfit on the CAL category, a relatively simpler classification, to compensate for their inability to handle more complex scenarios \citep{tirumala2022memorization, xu2021raise}. Models exhibiting this phenomenon can assign different weights to samples of different categories during training to reduce the model's preference for a particular category. This can be achieved by adjusting the weights in the loss (\textit{e.g.}, Focal Loss \& AdaFocal) \citep{li2022generalized,ghosh2022adafocal}.

\textbf{Finding \#3: MLLMs with strong overall performance tend to handle STEP easier than CATE.} From Figures \ref{fig:cate_performance_bar_top10} and \ref{fig:step_performance_bar_top10}, the best open-source MLLMs, such as InternVL2-76B, and the best closed-source MLLMs, like GPT-4o, exhibit a tendency where their STEP performance surpasses that of CATE. This trend holds even for human performance, where accuracy on STEP is higher (69.8\%) compared to recall on CATE (60.7\%). The reason for this disparity is likely that identifying the error step is inherently easier, as it involves localizing a specific point of failure. On the other hand, categorizing the error requires more complex reasoning and contextual understanding to classify the nature of the error, which adds difficulty. This mirrors the settings in object detection, where localization (\textit{i.e.,} predicting where an object is) is relatively simpler than classification (\textit{i.e.,} predicting what an object is) \citep{zou2023object,jiao2021new}. To improve the performance of error categorization tasks, MLLMs need to better understand the relationship between the problem itself and the steps where errors occur. Thus, modeling this part of the relationship can be a focus in the design of training data \citep{shi2024continual,song2025prmbench}.

\textbf{Finding \#4: CAL is the easiest category for MLLMs, while KNOW is the most difficult.} CAL is the category with the highest F1 performance among most MLLMs, which could be attributed to the structured and deterministic nature of calculations, where errors often result in clear, quantifiable deviations from expected outcomes, making them more straightforward to detect \citep{lewkowycz2022solving,kojima2022large}. Conversely, KNOW stands out as the most challenging category, suggesting that MLLMs struggle significantly with tasks requiring deep factual understanding and contextual reasoning. The complexity of knowledge errors likely stems from the need for comprehensive domain expertise, which current MLLMs may not fully encapsulate yet. 

\textbf{Finding \#5: MLLMs still have a gap to close to reach human-level intelligence in error detection.} Human performance significantly outperforms the best MLLMs in both the STEP and CATE tasks, with overall performance of 69.8\% and 60.7\% respectively, compared to the highest MLLM scores of 55.1\% and 53.1\%. Notably, the detection of VIS by humans is markedly superior to the best MLLMs, with a difference of nearly 20\%. This substantial lead may be attributed to the sophisticated pattern recognition inherent to human visual processing \citep{doerig2022visual}, which MLLMs, despite their advancements, have yet to fully emulate. Besides, it is interesting to note that human performance in REAS detection is lower than all closed-source MLLMs but higher than almost all open-source MLLMs. 

\begin{figure}[ht]
    \centering
    \includegraphics[width=1\linewidth]{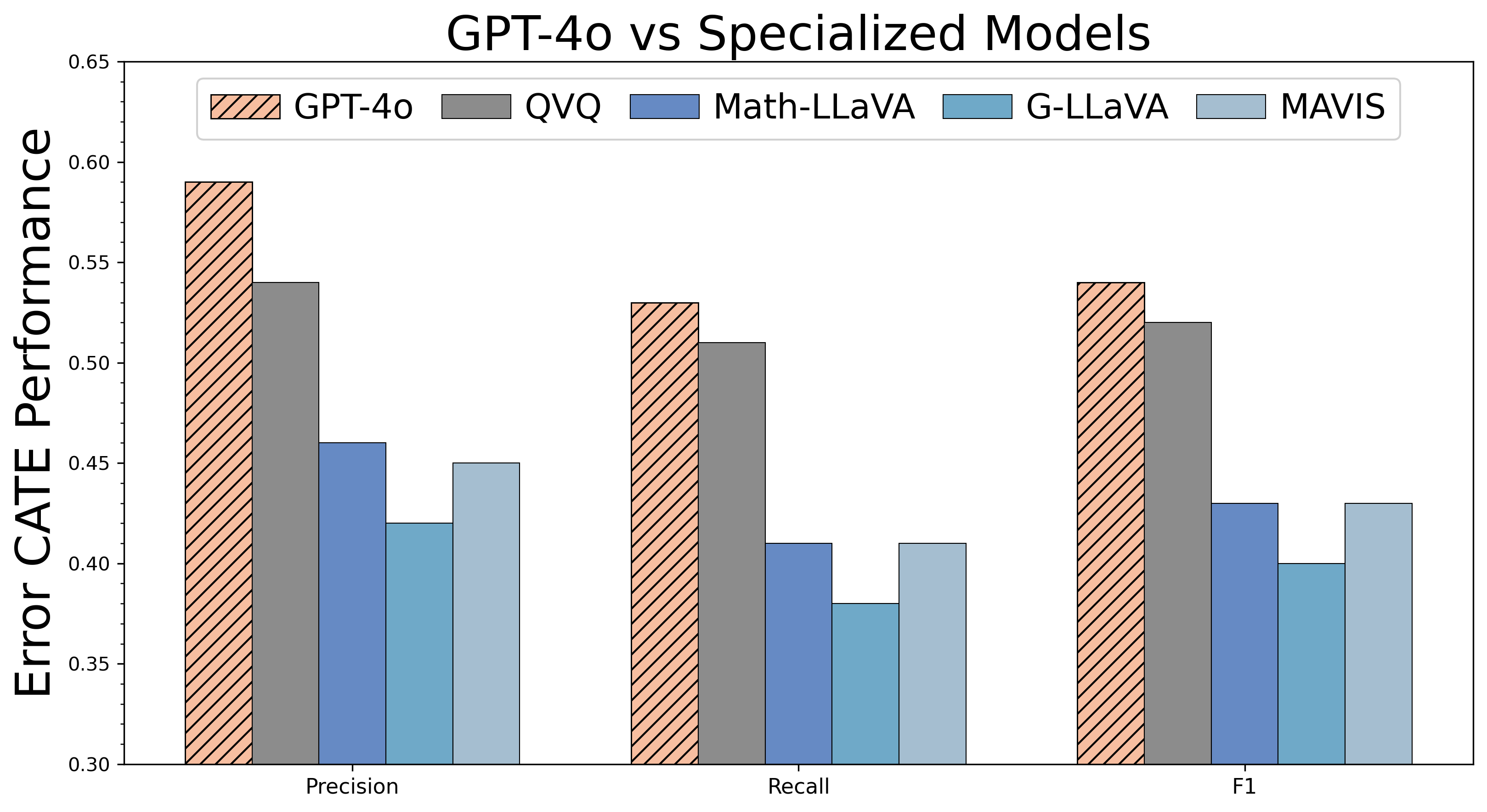}
    \caption{Performance comparison between GPT-4o vs multimodal reasoning-/math-specialized models.}
    \label{fig:specialized_models_performance}
    \vspace{-2mm}
\end{figure}

\textbf{Finding \#6: Specialized multimodal reasoning and math models underperform generalist models like GPT-4o.} As illustrated in Figure \ref{fig:specialized_models_performance}, we evaluate QVQ \cite{qvq-72b-preview} (a reasoning-enhanced variant of Qwen2-VL-72B) and find that it still lags behind GPT-4o. This suggests that specialized reasoning training alone does not guarantee superior performance in our task, which requires fine-grained multimodal error analysis beyond pure logical deduction. We also test three math-focused models \cite{zhang2024mavis,gao2023gllava,shi2024mathllava} to investigate whether problem-solving ability generalizes to our task. Their performance was weaker than general models, with G-LLaVA showing the lowest scores. We attribute this to its narrow geometric-focused training, which lacks diversity for our error categories.
\subsubsection{Visual Perception Analysis}
\label{sec:visual_analysis}

Due to the space limit, we discuss the visual perception analysis in Appendix \ref{app:visual_analysis}, and further analyze misclassification for each category and visual perception case study in Appendix \ref{app:confusion_matrix} and \ref{app:vis_bad_category_gpt}.

\subsubsection{Relation between Error Category and Error Step}
\label{sec:step percentage}
Due to the space limit, we discuss the relation between error category and error step in Appendix \ref{app:step_percentage}, and further analyze the phenomenon via cognitive load analysis in Appendix \ref{app:cognitive_load}.

\subsubsection{Scaling Analysis}
\label{sec:hyperparameter_analysis}

\begin{figure}[!t]
    \centering
    \includegraphics[width=0.8\linewidth]{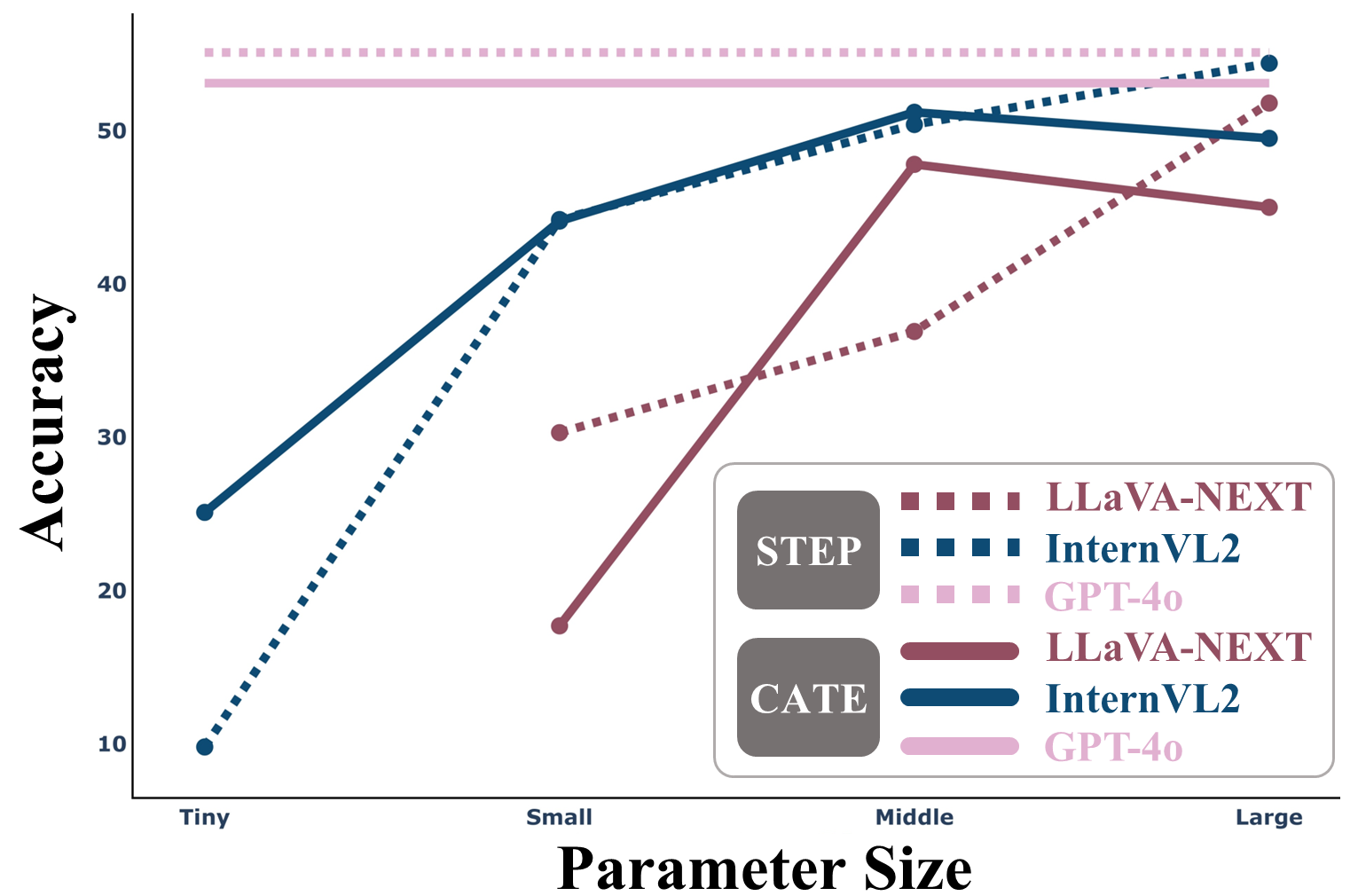}
    \caption{The accuracy of \textbf{STEP} and \textbf{CATE} of two representative MLLM series: \red{LLaVA-NEXT} and \blue{InternVL2}. We denote \textit{Tiny}, \textit{Small}, \textit{Middle}, \textit{Large} as the 2B, 8B, 26B, 76B for InternVL2 and None, 7B, 13B, 72B for LLaVA-NEXT.}
    \label{fig:scale_acc}
    \vspace{-4mm}
\end{figure}

\textbf{Finding \#1: The performance of MLLMs on STEP task increases with the scale of parameters.} We observe a phenomenon similar to the scaling law~\citep{kaplan2020scaling} in our experiments. As shown in Figure \ref{fig:scale_acc}, when the size of the InternVL2 model increases from Tiny to Huge, the accuracy of STEP task rises from 9.8\% to 54.4\%, showing an improvement of 44.6\%. Similarly, as the size of LLaVA-NEXT increases from Small to Large, its STEP accuracy also improves from 30.3\% to 51.8\%, indicating larger MLLMs exhibit greater reasoning ability in localizing error steps.\par
\textbf{Finding \#2: CATE task is relatively more difficult to improve through scaling.} While the accuracy of CATE shows a trend of improvement for both the InternVL2 and LLaVA-NEXT models as their size increases from Tiny (Small) to Middle, a slight decrease is also observed when model size reaches Large. We presume this is because CATE is a more challenging task compared to STEP, and merely increasing the model size without fine-tuning is insufficient for sustained improvement and may even introduce bias \citep{aghajanyan2023scaling,muennighoff2024scaling}.
\section{Conclusion}

In conclusion, we introduce \dataset, the first multimodal benchmark designed specifically for evaluating MLLMs's reasoning in mathematical error detection scenarios. By focusing on both \textit{error step identification} and \textit{error categorization}, \dataset bridges a critical research gap in assessing MLLMs’ capabilities in complex mathematical reasoning. The dataset's construction, based on real-world student interactions, ensures a robust evaluation framework that reflects genuine user needs. Extensive experimental analysis, comparing leading open-source and proprietary MLLMs, reveals significant challenges in error detection, highlighting the need for continued advancements in the multimodal reasoning domain towards Artificial General Intelligence. 

\section*{Limitations}

While \dataset provides a novel benchmark for multimodal mathematical error detection, we acknowledge certain limitations that also pave the way for future research:

\begin{itemize}
    \item \textbf{Dataset Scale:} Although our dataset of 2,500 instances is substantial for an initial benchmark and sourced from real-world interactions, the vast domain of K-12 mathematics encompasses an even wider array of problem types, visual representations, and nuanced student errors. Future work will focus on continuously expanding \dataset with more instances and greater diversity in problem structures and visual elements to ensure broader coverage.

    \item \textbf{Scope of Evaluated MLLMs:} Our study benchmarked a comprehensive set of over 20 contemporary MLLMs. Nevertheless, the field of MLLMs is evolving at an extremely rapid pace, with new architectures and larger models being released frequently. We aim to periodically update our benchmark results by evaluating new state-of-the-art MLLMs as they become available, ensuring \dataset remains a relevant and current tool.

    \item \textbf{Static Error Evaluation:} The benchmark assesses MLLMs on their ability to identify and categorize errors in pre-existing, static student solutions. It does not currently evaluate the models' interactive capabilities, such as guiding a student through error correction or generating pedagogical explanations for the identified errors. We envision future research building upon \dataset to explore these more dynamic and interactive educational applications of MLLMs.
\end{itemize}

\section*{Acknowledgments}
This work was supported by the Squirrel AI research fund; National Natural Science Foundation of China (Grant No.62506318); Guangdong Provincial Department of Education Project (Grant No.2024KQNCX028); Scientific Research Projects for the Higher-educational Institutions (Grant No.2024312096), Education Bureau of Guangzhou Municipality; Guangzhou-HKUST(GZ) Joint Funding Program (Grant No.2025A03J3957), Education Bureau of Guangzhou Municipality.


\bibliography{errorradar}

\clearpage
\newpage
\appendix
\hypersetup{linkcolor=black}
\etocdepthtag.toc{mtappendix}
\etocsettagdepth{mtchapter}{none}
\etocsettagdepth{mtappendix}{section}
\etocsettagdepth{mtappendix}{subsubsection}
\tableofcontents
\clearpage

{\large\textbf{Technical Appendices and Supplements}}

\section{Comparison with Relevant Benchmarks}
\label{app:bench_comprison}

The field of mathematical reasoning evaluation has seen a significant proliferation of benchmarks, each designed to probe the capabilities of LLMs and MLLMs. As illustrated in Table~\ref{tab:bench comparison}, these benchmarks can be broadly categorized.

A substantial portion of existing work has concentrated on text-based mathematical problems. Benchmarks such as TheoremQA~\citep{chen2023theoremqa}, MathBench~\citep{liu2024mathbench}, MR-GSM8K~\citep{zeng2024mr}, and SciEval~\citep{sun2024scieval} provide valuable resources for assessing the pure linguistic and logical reasoning abilities of LLMs on mathematical tasks. Recognizing the inherently multimodal nature of many real-world math problems, the community has also developed numerous benchmarks that integrate visual information. Datasets like CMMaTH~\citep{li2024cmmath}, MathScape~\citep{zhou2024mathscape}, MATH-V~\citep{wang2024measuring}, MathVista~\citep{lu2023mathvista}, and MathVerse~\citep{zhang2024mathverse} challenge MLLMs to solve problems by jointly interpreting textual and visual inputs (See more discussion on these benchmarks in Appendix \ref{app:math_benchmarks}). However, a common thread among these benchmarks is their primary focus on evaluating a model's \textbf{problem-solving ability}—that is, their capacity to generate a correct final answer. More recently, the task of error analysis has begun to emerge. For instance, the EIC benchmark~\citep{li2024evaluating} made a notable contribution by introducing tasks for error identification and correction. While this represents a critical step toward evaluating deeper reasoning, EIC is limited to a \textit{text-only} modality.

In this context, \dataset is uniquely positioned to fill two critical gaps in the current evaluation landscape. First, it is the \textbf{first benchmark designed to specifically evaluate multimodal error detection}. By requiring models to analyze reasoning steps in a context that includes both text and images, it presents a more complex and realistic challenge that moves beyond simple problem-solving. Second, and most distinctively, \dataset is built upon \textbf{real student answers}. Unlike benchmarks that rely on synthetically generated data or correct solutions, our dataset captures the authentic, often nuanced errors that human learners make. This provides a more robust and educationally relevant setting to test the fine-grained diagnostic capabilities of MLLMs.

In summary, while previous benchmarks have laid essential groundwork for evaluating mathematical problem-solving, \dataset introduces a novel, more demanding paradigm focused on multimodal error diagnostics with real-world data, thereby offering a more comprehensive assessment of complex mathematical reasoning.

\section{Impact Statement}
\label{app:impact_statement}

\begin{figure}[t!]
    \centering
        \begin{table}[H]
            \centering
            \begin{adjustbox}{width=0.45\textwidth}
            \begin{tabular}{ccccccc}
                \toprule
                \textbf{Benchmarks} & \textbf{Venue} & \textbf{Modality} & \textbf{Student Ans.} & \textbf{Error Det.} \\
                \midrule
                 \multicolumn{1}{l}{TheoremQA \citep{chen2023theoremqa}}
                & EMNLP & $T$ & - & -  \\ 
                \multicolumn{1}{l}{MathBench \citep{liu2024mathbench}}
                & ACL & $T$ & - & -  \\ 
                \multicolumn{1}{l}{MR-GSM8K \citep{zeng2024mr}}
                & ICLR & $T$  & - & -  \\ 
                \multicolumn{1}{l}{SciEval \citep{sun2024scieval}}
                & AAAI & $T$  & - & - \\ 
                \multicolumn{1}{l}{EIC \citep{li2024evaluating}}
                & ACL Finding & $T$  & - & \checkmark   \\ 
                \multicolumn{1}{l}{CMMaTH \citep{li2024cmmath}}
                & COLING & $T,I$ & - & -  \\ 
                \multicolumn{1}{l}{MathScape \citep{zhou2024mathscape}}
                & arXiv & $T,I$ & - & -  \\ 
                \multicolumn{1}{l}{MATH-V \citep{wang2024measuring}}
                & NeurIPS & $T,I$ & - & -  \\ 
                \multicolumn{1}{l}{Olympiadbench \citep{he2024olympiadbench}}
                & ACL & $T,I$ & - & -  \\ 
                \multicolumn{1}{l}{QRData \citep{liu2024llms}}
                & ACL & $T,I$ & - & -  \\ 
                \multicolumn{1}{l}{IsoBench \citep{fu2024isobench}}
                & COLM & $T,I$ & - & -  \\ 
                \multicolumn{1}{l}{SciBench \citep{wang2023scibench}}
                & ICML & $T,I$ & - & -  \\
                \multicolumn{1}{l}{EMMA \citep{hao2025can}}
                & ICML & $T,I$ & - & -  \\
                \multicolumn{1}{l}{MathCheck \citep{zhou2024mathcheck}}
                & ICLR & $T,I$ & - & -   \\ 
                \multicolumn{1}{l}{DynaMath \citep{zou2024dynamath}}
                & ICLR & $T,I$ & - & -   \\ 
                \multicolumn{1}{l}{MathVista \citep{lu2023mathvista}}
                & ICLR & $T,I$ & - & -   \\
                \multicolumn{1}{l}{Math-Vision \citep{wang2024mathvision}}
                & NeurIPS & $T,I$ & - & -   \\
                \multicolumn{1}{l}{MV-MATH \citep{wang2025mvmath}}
                & CVPR & $T,I$  & - & - \\ 
                \multicolumn{1}{l}{CMM-MATH \citep{liu2025cmm}}
                & ACM MM & $T,I$  & - & - \\ 
                \multicolumn{1}{l}{MathVerse \citep{zhang2024mathverse}}
                & ECCV & $T,I$  & - & - \\ 
                \rowcolor{gray!40} \multicolumn{1}{l}{\dataset (Ours)}
                & - & $T,I$  & \checkmark & \checkmark \\ 
                \bottomrule
            \end{tabular}
            \end{adjustbox}
            \caption{Comparison between our proposed \dataset benchmark vs. its relevant LLM-based mathematical reasoning benchmarks or datasets. Under the column of \textbf{\textit{Modality}}, the letters $T$ and $I$ represent text and image, respectively. The column labeled as \textbf{\textit{Student Ans.}} indicates whether the dataset contains real student data (\textit{i.e.}, students' incorrect answers); the column labeled as \textbf{\textit{Error Det.}} represents whether error detection task is included.}
            \label{tab:bench comparison detail}
            \vspace{-2.5em}
        \end{table}    
\end{figure}

The introduction of the \dataset benchmark and the formalization of the multimodal error detection task represent significant strides in enhancing the complex reasoning abilities of MLLMs. We will release the whole dataset to the community upon acceptance, and we will be committed to refining and scaling up the dateset for further research. Its broader impact can be seen in several key areas:

First, the ability to detect and categorize errors in mathematical reasoning is critical for improving the efficacy of AI in educational settings. By using real-world data sourced from student interactions, \dataset provides invaluable insights into the cognitive and error patterns of learners. These insights can inform personalized learning interventions, helping to identify not only where students struggle but also why they do so. 

Second, while the use of AI in educational settings presents substantial opportunities, it also necessitates a careful consideration of fairness, bias, and transparency. By incorporating human expert evaluation and real-world student data, \dataset helps to mitigate risks of biased AI assessments, ensuring that MLLMs are held to high standards of accuracy across diverse error categories. However, as these models continue to evolve, it is crucial that they remain accessible and equitable across different educational contexts.

Last, the success of multimodal AI in addressing complex mathematical reasoning tasks could have a transformative effect on education at large, extending beyond traditional K-12 settings into higher education and lifelong learning. The \dataset benchmark sets the stage for further research into error detection across diverse disciplines, such as physics, economics, and even coding, where MLLMs could be used to enhance learning outcomes through improved diagnostic and instructional capabilities.

\section{Task Clarification}
\label{app:task_clarification}

\subsection{How \dataset Contribute to Complex Multimodal Math Reasoning}
\label{app:how_contribute}
While our primary focus in this paper was to formally introduce the task, construct the benchmark, and establish baseline performances (as is typical for a benchmark paper), we argue that evaluating and understanding error detection directly contributes to the goal of enhancing complex mathematical reasoning in several crucial ways:

\begin{itemize}
    \item \textbf{Error Detection as a Diagnostic Tool for Deeper Reasoning Failures:} Standard problem-solving benchmarks typically evaluate only final answer accuracy, with limited insight into why a model fails. \dataset offers a more granular diagnostic capability, via identifying where the reasoning breaks down and what kind of mistake was made.
    \item \textbf{Highlighting Specific Weaknesses for Targeted Improvement:} Our extensive experiments (See Section \ref{sec:experiment}) reveal distinct error patterns across MLLMs:
    \begin{itemize}
        \item Weaker open-source models often exhibit a bias towards predicting CAL, suggesting they oversimplify complex issues.
        \item KNOW is challenging for most models.
        \item Top models like GPT-4o struggle with visual perception compared to humans. These findings pinpoint specific areas (\textit{e.g.,} visual grounding, knowledge integration) where models need improvement.
        \item We explicitly provide \textbf{Actionable Suggestions} in Appendix \ref{app:actionable suggestions}, directly linking the benchmark's insights to pathways for enhancing MLLM capabilities.
    \end{itemize}
    \item \textbf{Error Detection as a Proxy for Robust Reasoning:} The ability to correctly identify an error in a given reasoning chain requires a deep understanding of the correct reasoning process itself. A model that can reliably perform error detection inherently possesses a more sophisticated grasp of mathematical logic, visual interpretation, and knowledge application. Therefore, success in error categorization signals a higher level of metacognitive-like reasoning.
\end{itemize}

\subsection{Comparison between \dataset Setting and General LLM-as-Judge}
\label{app:comparison_llm_as_judge}
To clearly illustrate the distinctions, we present the following comparison:

\begin{table*}[htbp]
    \centering
    \caption{Comparison between \dataset and general LLM-as-Judge paradigm.}
    \label{tab:errorradar_vs_llm_judge_comparison}
    \small 
    \begin{adjustbox}{max width=\textwidth}
    \begin{tabular}{@{} l c c @{}}
        \toprule
        \textbf{Dimension} & \textbf{\dataset} & \textbf{General LLM-as-Judge} \\
        \midrule
        \textbf{Input Data} & 
        \makecell{Real student incorrect step-by-step \\ solutions \& problem context} & 
        \makecell{Often model-generated \\ outputs} \\
        \midrule
        \textbf{Core Task} & 
        \makecell{Specific \& \\ Fine-grained} & 
        \makecell{General \& \\ Variable} \\
        \midrule
        \textbf{Output Granularity} & 
        \makecell{Precise \& \\ Structured} & 
        \makecell{Score, ranking, \\ etc.} \\
        \midrule
        \textbf{Evaluation Focus} & 
        \makecell{Diagnostic \\ Accuracy} & 
        \makecell{Overall correctness, \\ coherence, etc.} \\
        \midrule
        \textbf{Context/Purpose} & 
        \makecell{Educational \\ Diagnostics} & 
        \makecell{Model \\ evaluation} \\
        \midrule
        \textbf{Error Taxonomy} & 
        \makecell{Predefined \& Educationally \\ Relevant} & 
        \makecell{Usually no fixed, \\ fine-grained error taxonomy} \\
        \bottomrule
    \end{tabular}
    \end{adjustbox}
\end{table*}

\vspace{\medskipamount} 

\noindent 
Therefore, while identifying wrong steps is a facet of some LLM-as-Judge applications, ErrorRadar defines a significantly \textbf{more specific}, \textbf{complex}, and \textbf{context-grounded} set of tasks.

\subsection{Relationship between \dataset Setting and Multimodal Math Problem-Solving}
\label{app:relation_with_problem_solving}
As shown in Section \ref{sec:main_result} Finding \#6, results offer compelling evidence regarding this relationship:

\textbf{Evidence of Relationship}: The fact that these math-specialized models can perform the \dataset tasks demonstrates a fundamental relationship. Both problem-solving and error detection operate within the multimodal mathematics domain. They require core capabilities like understanding mathematical notation and terminology presented textually.

\textbf{Evidence of Distinction}: The performance of these specialized models on \dataset was only moderate. This performance gap provides evidence that the capabilities required by \dataset are distinct from, and not merely a subset of, problem-solving ability. The math-specialized models’ training optimizes for generating correct solution paths but may not sufficiently equip them to diagnose errors in incorrect paths.

While our benchmark primarily focuses on evaluating MLLMs’ ability to detect errors in student-provided reasoning chains (a critical task for educational applications), we acknowledge the importance of assessing \textit{whether MLLMs can independently solve the same problems correctly}. To address this, we conducted additional experiments: GPT-4o achieved 82\% accuracy in solving problems independently, significantly higher than its error detection performance. This aligns with prior work (\textit{e.g.,} MathVista \cite{lu2023mathvista}) showing that MLLMs excel in direct problem-solving but struggle with error analysis.

\section{More Related Work}
\label{app:more_related_work}
\subsection{Mathematical Reasoning Benchmarks}
\label{app:math_benchmarks}

The landscape of mathematical reasoning benchmarks has evolved significantly, \textbf{initially focusing on textual problem-solving and gradually incorporating multimodal inputs and more complex reasoning tasks.} Early efforts in benchmarking, such as MathQA \cite{amini2019mathqa} and GSM8K \cite{cobbe2021training}, along with the widely used MATH dataset \cite{hendrycks2021measuring}, primarily assessed models on arithmetic and word problems presented in a text-only format. These were instrumental in the pre-LLM and early LLM era. As models grew more sophisticated, benchmarks like LILA \cite{mishra2022lila} continued to expand the scope of text-based mathematical reasoning. However, recognizing that many real-world mathematical problems inherently involve visual components, the research community has increasingly developed multimodal benchmarks. Prominent examples include MathVista \cite{lu2023mathvista}, which evaluates reasoning in visual contexts, MathVerse \cite{zhang2024mathverse} focusing on whether MLLMs truly understand diagrams, and MMMU \cite{yue2024mmmu} for massive multi-discipline multimodal understanding. Specific multimodal domains like geometry have also seen dedicated benchmarks such as GeoQA \cite{chen2021geoqa}, GeomVerse \cite{kazemi2023geomverse}, GeoEval \cite{zhang2024geoeval}, MATH-Vision \cite{wang2024measuring}, and DynaMath \cite{zou2024dynamath} which introduces dynamic visual elements. Efforts to diversify language coverage in multimodal settings are also emerging, with datasets like CMM-Math \cite{li2024cmmath} for Chinese.

Beyond the transition to multimodality, benchmark tasks have expanded \textbf{from straightforward problem-solving to encompass more intricate aspects of mathematical reasoning}. While problem-solving remains a dominant task, as seen in competition-level benchmarks like Omni-MATH \cite{gao2024omni}, CHAMP \cite{mao2024champ}, and PutnamBench \cite{tsoukalas2024putnambench}, there’s a growing focus on higher-order thinking. This includes mathematical proving, evaluated by datasets such as TheoremQA \cite{chen2023theoremqa} and the IMO-AG-30 subset used with AlphaGeometry \cite{trinh2024solving}. A particularly significant development is the increased attention to error analysis. Benchmarks like EIC-Math \cite{li2024evaluating} and Mathador-LM \cite{kurtic2024mathador} focus on error identification and, in some cases, correction within textual solutions. FaultyMath \cite{rahman2024blind} evaluates logical integrity on faulty problems. The complexity of evaluation is further pushed by benchmarks addressing multi-turn interactions like MathChat \cite{liang2024mathchat}, model robustness through adversarial examples (GSM-PLUS by \cite{li2024gsmplus}), and specialized quantitative reasoning with data (QRData by \cite{liu2024llms}) or mathematical modeling (Mamo by \cite{huang2024mamo}). These advancements reflect a drive towards more comprehensive and challenging evaluations of (M)LLMs’ mathematical reasoning capabilities \cite{yan2024survey,yan2025mathagent}.

\subsection{Math-Specific MLLMs}
\label{app:math_specific_mllms}
The evolution of AI in mathematical reasoning has significantly advanced with the advent of Math-Specific MLLMs (Math-MLLMs), which are engineered to interpret and resolve mathematical problems incorporating both textual and crucial visual elements like diagrams, graphs, and geometric figures \cite{yan2024survey}. Among the notable developments in this domain is MathGLM-Vision \cite{yang2024mathglm}, a model explicitly designed to integrate visual information for solving mathematical problems. Similarly, Math-LLaVA \cite{shi2024mathllava} leverages fine-tuning on an extensive dataset of 360K high-quality multimodal math question-answer pairs (MathV360K) to directly enhance its multimodal mathematical reasoning capabilities. Addressing the challenge of data scarcity, MAVIS \cite{zhang2024mavis} features an automatic data generation engine and employs instruction fine-tuning to teach models problem decomposition. While primarily text-based, models such as Skywork-Math \cite{zeng2024skywork} and Qwen2.5-Math \cite{bai2025qwen25vl} are also adapting to support multimodal mathematical settings. Further contributions include Math-PUMA \cite{zhuang2025mathpuma} with its progressive upward multimodal alignment strategy, and InfiMM-Math \cite{han2024infimm}, which achieves strong performance through training on a large-scale, LLM-validated multimodal dataset. Methodological innovations are also prominent, such as Visual Sketchpad \cite{hu2024visual} enabling MLLMs to generate intermediate sketches, G-LLaVA \cite{gao2023gllava} focusing on geometry, STIC \cite{deng2024stic} employing self-training for visual comprehension, and VCAR \cite{jia2024vcar} emphasizing visual-centric supervision. These collective efforts highlight a strong push towards integrating visual perception with textual understanding, though challenges in advanced visual reasoning and the need for diverse, large-scale datasets remain critical for future progress \cite{yan2024survey,yan2025position}.

\section{More Multimodal Question Examples}
\label{app:example}

\begin{figure*}[th!]
  \centering
  \includegraphics[width=\textwidth]{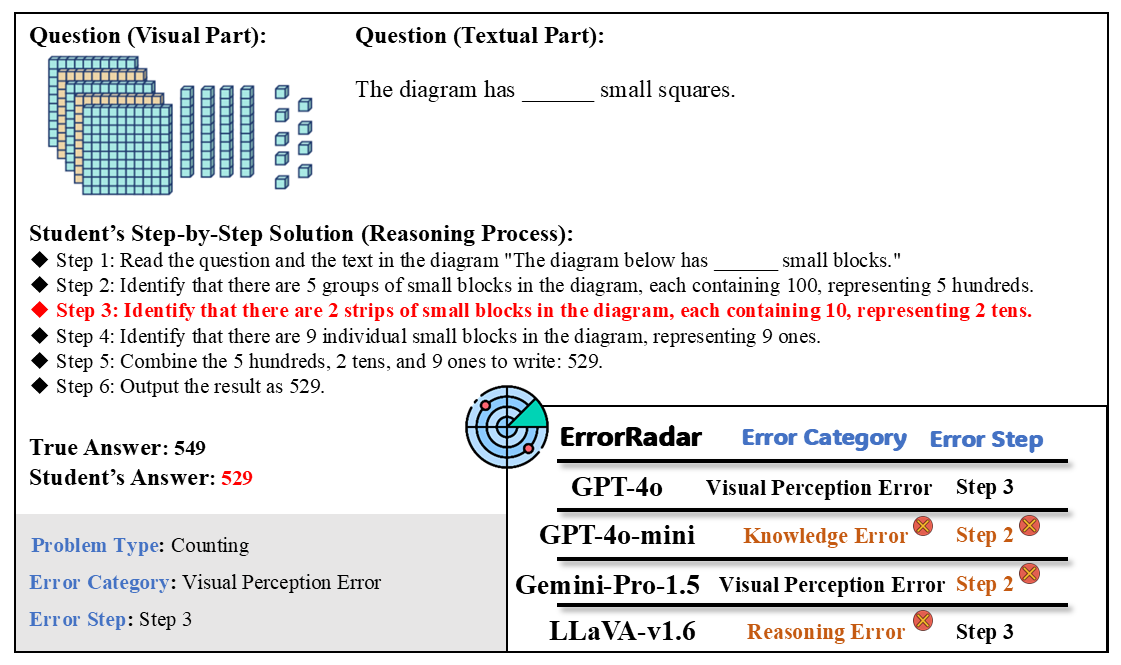}
  \caption{Multimodal mathematical example one (type: counting) from \dataset dataset.}
\label{fig:case1}
\end{figure*}

\begin{figure*}[th!]
  \centering
  \includegraphics[width=\textwidth]{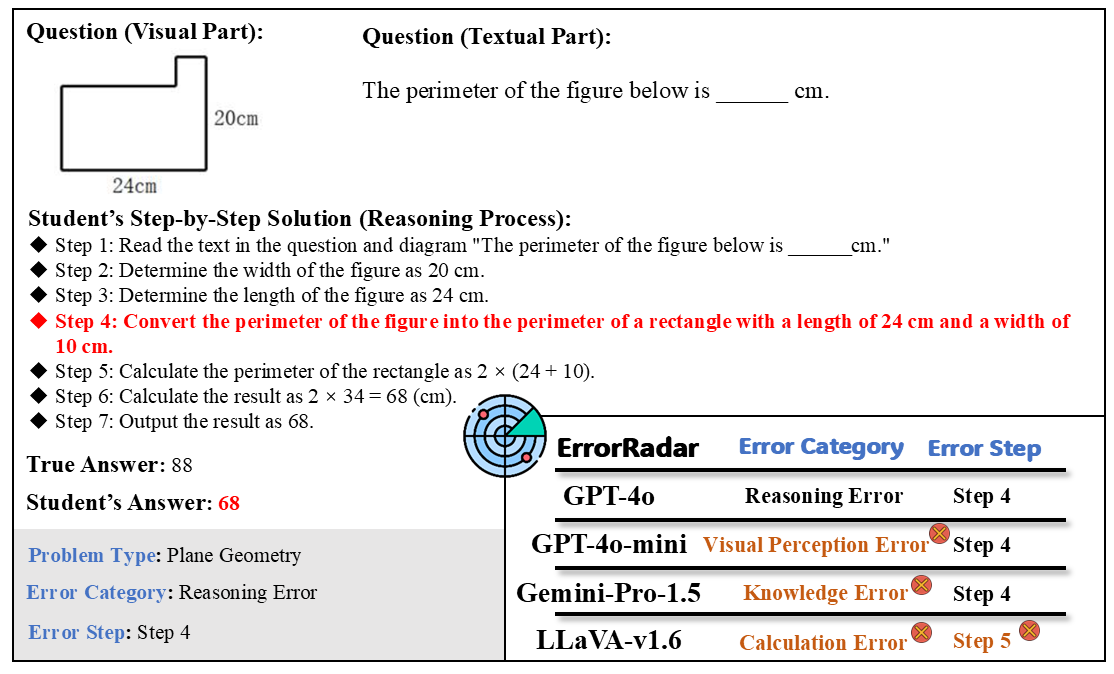}
  \caption{Multimodal mathematical example two (type: plane geometry) from \dataset dataset.}
\label{fig:case2}
\end{figure*}

\begin{figure*}[th!]
  \centering
  \includegraphics[width=\textwidth]{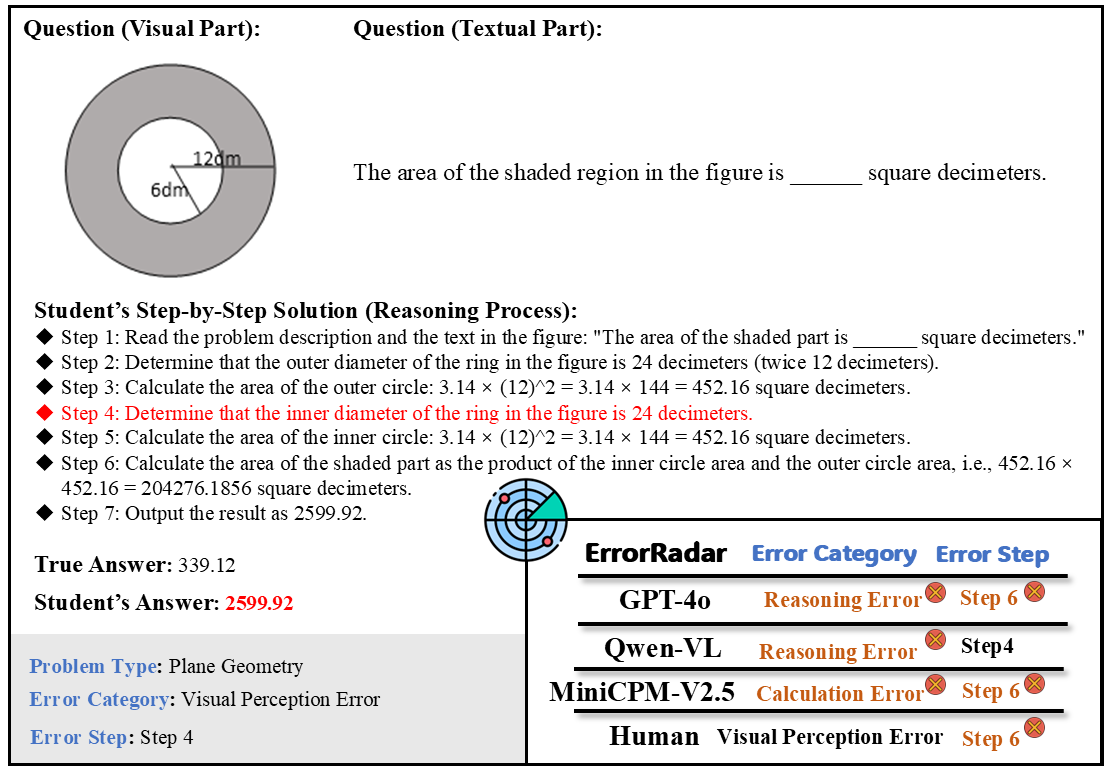}
  \caption{Multimodal mathematical example three (type: plane geometry) from \dataset dataset.}
\label{fig:case3}
\end{figure*}

\begin{figure*}[th!]
  \centering
  \includegraphics[width=\textwidth]{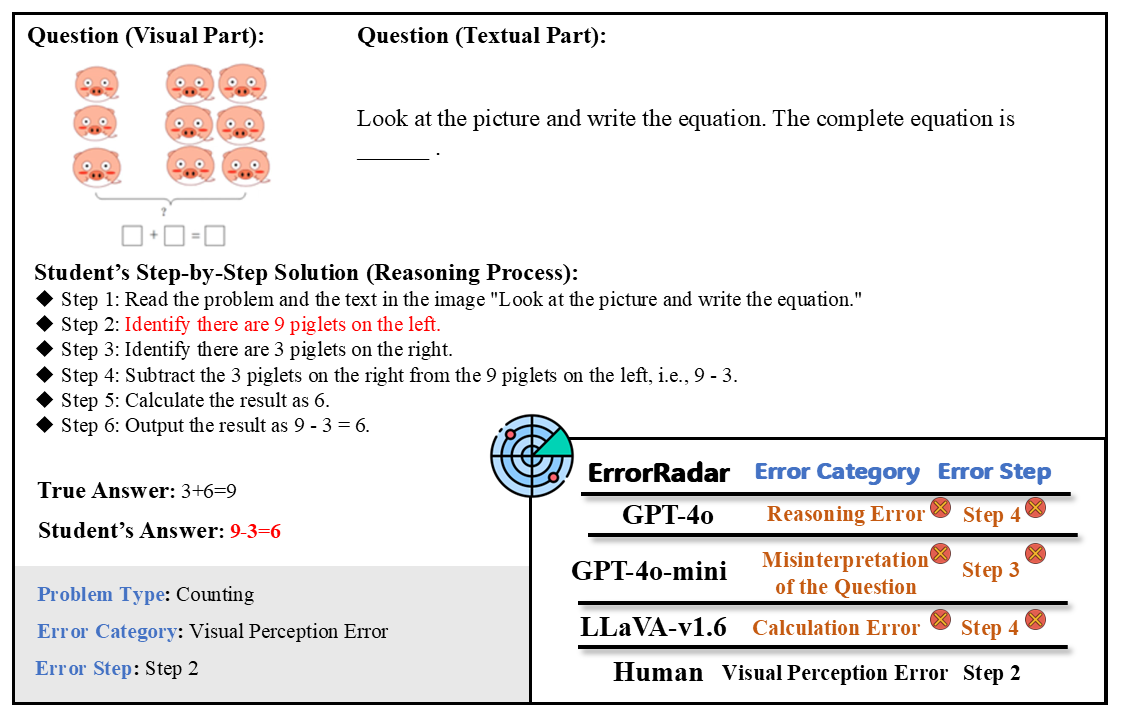}
  \caption{Multimodal mathematical example four (type: counting) from \dataset dataset.}
\label{fig:case4}
\end{figure*}

\begin{figure*}[th!]
  \centering
  \includegraphics[width=\textwidth]{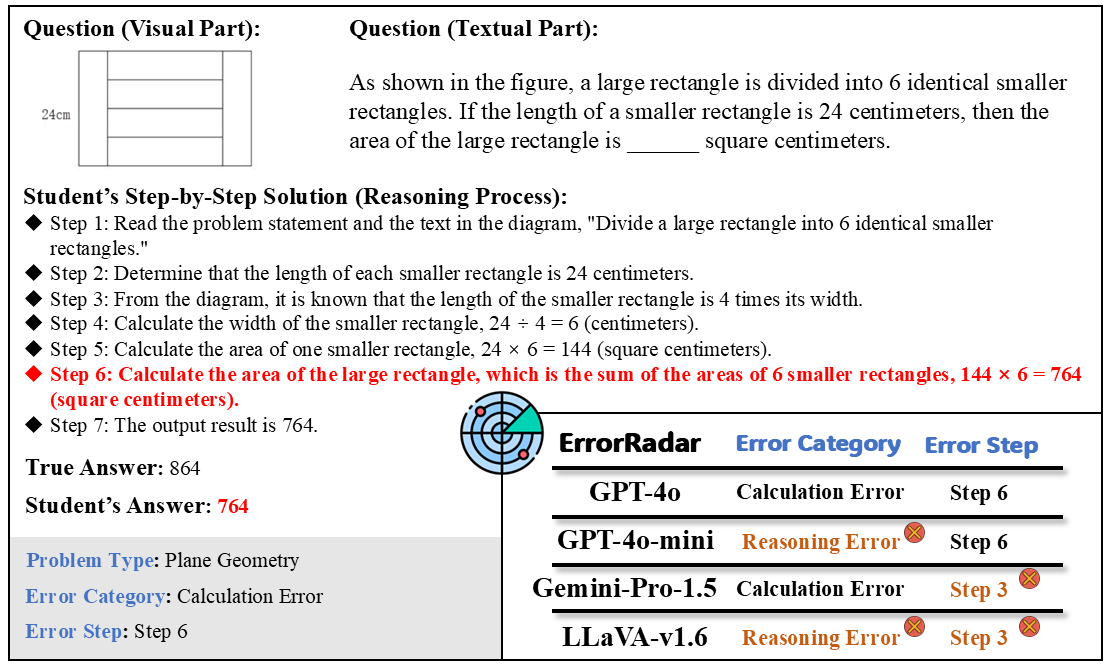}
  \caption{Multimodal mathematical example five (type: plane geometry) from \dataset dataset.}
\label{fig:case5}
\end{figure*}

In this section, you can refer to Figures \ref{fig:case1}, \ref{fig:case2}, \ref{fig:case3}, \ref{fig:case4}, and \ref{fig:case5} for more concrete examples from our \dataset dataset.

\section{Additional Dataset Details}
\label{app:dataset}
\subsection{Annotation Details}
\label{app:annotation detail}
To ensure the quality and relevance of the \dataset dataset for error detection tasks, we employed a rigorous manual annotation process, involving professional educational experts as annotators. This section outlines the details of the annotation procedure, focusing on how the data was enriched with step-by-step reasoning processes, identification of erroneous steps, and error categorization.

\textbf{Annotator Selection and Training.} Given the complexity of the task, we recruited a group of ten annotators with specialized knowledge in educational theory and mathematics, particularly in K-12 pedagogy. These annotators were trained extensively to familiarize themselves with the structure and expectations of the task. The training covered the specifics of multimodal problem-solving in mathematics, typical student error patterns, and the need for precise identification of reasoning steps that led to incorrect answers. The annotators were also briefed on using the provided tools and the quality assurance process.

\textbf{Annotation Process.} Each mathematical problem in the dataset was annotated with a step-by-step reasoning process, capturing both correct and incorrect approaches to problem-solving. Annotators were provided with:

\begin{enumerate}
    \item The original question stem (comprising both text and image components).
    \item The student's most frequent incorrect answer.
    \item The correct answer to the question.
    \item The pedagogical analysis of the correct reasoning process, prepared by educational experts.
\end{enumerate}

Based on these inputs, annotators were tasked with:

\begin{enumerate}
    \item \textbf{Step-by-Step Reasoning Annotation}: For each problem, annotators mapped out the logical steps that students should ideally follow to arrive at the correct answer. This involved identifying key stages in the problem-solving process, such as formula application, arithmetic operations, or logical deductions.
    \item \textbf{Error Step Identification}: For problems where students provided incorrect answers, annotators identified the exact steps where the reasoning went wrong. These error steps were explicitly marked and linked to the incorrect responses, ensuring that they could be traced back to specific problem-solving mistakes.
    \item \textbf{Error Categorization}: Once the erroneous step was identified, annotators assigned an appropriate error category based on a predefined schema. These categories included common types of errors such as misinterpretation of the question (More details can be seen in Section \ref{sec:task_definition}). The categorization was designed to align with known student error patterns in mathematical learning.
\end{enumerate}

\textbf{Quality Control and Cross-Validation.} To ensure annotation accuracy and consistency, each problem underwent two rounds of cross-checking:

\begin{enumerate}
    \item \textbf{First Round of Cross-Validation}: After the initial annotation, another annotator independently reviewed the annotations. Any discrepancies between the first and second annotators were flagged for further analysis.
    \item \textbf{Second Round of Cross-Validation}: In the second round, if the errors or discrepancies persisted, the problem was escalated to a senior educational expert who acted as the annotation lead. The annotation lead adjudicated these contentious cases, ensuring that the final decision was both pedagogically sound and aligned with the dataset’s goals.
\end{enumerate}

\textbf{Dataset Refinement.} Throughout the annotation process, we worked closely with the educational organization from which the dataset originated. This collaboration ensured that the annotations were not only reliable but also adhered to the standards of the organization’s question bank. Additionally, ongoing feedback and updates from the organization helped refine the dataset, making it more accurate and relevant for multimodal error detection tasks.

\textbf{Annotation Duration and Effort.} The annotation process for the \dataset dataset spanned over a period of at least two months. During this time, the annotators, comprised of both professional educational experts and domain specialists, worked meticulously through several stages of preparation, annotation, and validation. Each annotator dedicated significant time to understanding the dataset, reviewing the provided pedagogical analyses, and applying their domain knowledge to identify and categorize errors. The first phase, involving step-by-step reasoning annotation, took approximately six weeks, while the subsequent cross-validation and quality control efforts accounted for the remaining two weeks. Given the complexity of the tasks and the necessity for high precision, the team’s sustained efforts ensured that the final dataset was of the highest quality.

By incorporating these annotations, \dataset provides a robust foundation for studying student errors in mathematical reasoning and enables the development of advanced models for error detection and correction.

\subsection{Details of Handling Inconsistent Annotations}
\label{app:annotation inconsistency handling}
To ensure the quality and reliability of our dataset for the multimodal mathematical error detection task, we established a systematic approach to resolve annotation inconsistencies. This process balances annotator independence with rigorous quality control, ensuring that the dataset is both accurate and representative.

\subsubsection{Annotation Agreement Principles}
\begin{enumerate}
    \item \textbf{Guided Consensus}: Annotations must align with clear, predefined guidelines covering the five error categories. Annotators are trained extensively to reduce subjective biases.
    
    \item \textbf{Cross-Checking and Agreement Threshold}: Each instance is annotated by at least three annotators. Disagreements are flagged for further review.

    \item \textbf{Systematic Review Process}: For inconsistent cases, a multi-step resolution process is applied:
    \begin{enumerate}
        \item \textbf{Initial Review}: Annotators discuss disagreements, referencing annotation guidelines and the specific problem context.
        \item \textbf{Expert Arbitration}: For unresolved cases, a domain expert (e.g., an educational professional) reviews and finalizes the annotation.
        \item \textbf{Consensus-Driven Decisions}: When possible, annotations are harmonized based on majority opinion or shared agreement after discussions.
    \end{enumerate}
\end{enumerate}

\subsubsection{Case Resolution Framework}

\paragraph{Case Example 1: Visual Perception vs. Reasoning Error}
\begin{itemize}
    \item \textbf{Example}: A problem presents a bar chart requiring students to determine the highest value. A student misidentifies the tallest bar and selects the wrong answer.
    \begin{itemize}
        \item Annotator A labels this as a Visual Perception Error, arguing the mistake stems from misreading the chart.
        \item Annotator B classifies it as a Reasoning Error, interpreting the mistake as a failure to compare values logically.
    \end{itemize}
    \item \textbf{Resolution: Annotators revisit the problem:}
    \begin{itemize}
        \item If evidence shows the student misunderstood the chart format (e.g., interpreting height as quantity but misjudging due to poor visualization), it is classified as a Visual Perception Error.
        \item If the student correctly interprets the chart but misapplies logical comparisons (e.g., failing to compare values explicitly), it is categorized as a Reasoning Error.
    \end{itemize}
    For persistent disagreement, an expert examines the student’s work, including any notes or intermediate steps, to determine the correct annotation.
\end{itemize}

\paragraph{Case Example 2: Knowledge vs. Misinterpretation of the Question}
\begin{itemize}
    \item \textbf{Example}: A problem asks for the perimeter of a rectangle, but the student calculates the area instead.
    \begin{itemize}
        \item Annotator A identifies this as a Knowledge Error, attributing the mistake to a lack of understanding of perimeter concepts.
        \item Annotator B labels it as a Misinterpretation of the Question, asserting that the student misunderstood what was being asked.
    \end{itemize}
    \item \textbf{Resolution}: 
    \begin{itemize}
        \item Did the student’s work demonstrate understanding of the concept but apply it incorrectly (Misinterpretation of the Question)?
        \item Did the mistake reveal a fundamental gap in knowledge about perimeter (Knowledge Error)?
    \end{itemize}
    If disagreement persists, the annotators consult the expert, who may analyze additional context (e.g., previous responses or annotations).
\end{itemize}

\subsubsection{Handling Irreconcilable Disagreements}
If discrepancies persist despite review and arbitration, the affected data points are excluded from the dataset. This strict policy prioritizes the overall quality and consistency of the dataset, ensuring that retained samples maintain high reliability.

\subsubsection{Monitoring and Feedback}
Periodic feedback sessions are conducted to recalibrate annotators and refine guidelines based on observed patterns of disagreement. This iterative approach minimizes future inconsistencies and enhances annotator alignment over time.

\subsection{Definition of Problem Type Category}
\label{app:problem type cate}
The \dataset dataset distinguishes five primary types of multimodal mathematical problems, each characterized by unique features:
\vspace{-0.6em}
\begin{itemize}[leftmargin=*]
    \item[\ding{79}]  \textbf{Plane Geometry Problems}: These involve two-dimensional shapes and figures, requiring knowledge of properties such as angles, lines, and polygons. Solving these problems often depends on understanding basic geometric principles and theorems about plane figures.
    \item[\ding{79}]  \textbf{Solid Geometry Problems}: In contrast to plane geometry, solid geometry involves three-dimensional objects, such as cubes, cylinders, and spheres. These problems require spatial visualization and understanding of volume, surface area, and the relationships between different three-dimensional shapes.
    \item[\ding{79}]  \textbf{Diagram-Based Problems}: These require analysis of provided visual information, such as graphs, charts, or diagrams, to solve mathematical queries. Interpreting visual data correctly is crucial, as these problems test the ability to extract and analyze quantitative information from visual aids.
    \item[\ding{79}]  \textbf{Algebra Problems}: Algebra problems focus on abstract symbols and variables to represent numbers and relationships. These include tasks like solving equations, manipulating algebraic expressions, and understanding functions. The problem-solving process typically involves logical reasoning and manipulation of mathematical symbols.
    \item[\ding{79}]  \textbf{Math Commonsense Questions}: These encompass a variety of problem types, including time judgment, direction judgment, counting, and pattern recognition. Unlike the other categories, math commonsense challenges rely on everyday mathematical reasoning and problem-solving strategies that do not necessarily require formal mathematical knowledge, testing intuitive understanding rather than procedural skills.
\end{itemize}
These problem types highlight the \dataset dataset's diverse nature, with each category presenting distinct challenges and requiring specific reasoning abilities.

\subsection{Development and Validation Process of Error Category}
\label{app:error category finalization process}
\subsubsection*{1. Cross-Team Collaboration to Align Task Needs}
The process began with close collaboration between the research team and the education team to ensure that the error categories aligned with the unique requirements of the multimodal math error detection task. The research team provided insights into the task’s technical objectives, focusing on precision and comprehensive error coverage. Simultaneously, the education team contributed their understanding of real-world educational scenarios, emphasizing the practical relevance and applicability of the error taxonomy to students’ and teachers’ needs.

\textbf{Key Outcomes:}
\begin{itemize}
    \item Initial consensus that the categories must address both multimodal challenges and real-life classroom scenarios.
    \item Recognition of the need to balance academic rigor with user-friendly categorization.
\end{itemize}

\subsubsection*{2. Benchmark Survey and Focus Analysis}
The research team conducted an extensive survey of representative benchmarks, focusing on error analysis frameworks in existing datasets. Examples included studies on problem-solving steps in educational AI and cognitive error modeling in multimodal tasks. The aim was to identify gaps in current frameworks and understand how existing taxonomies handle errors specific to visual, textual, and logical reasoning elements.

\textbf{Key Outcomes:}
\begin{itemize}
    \item Identification of inadequacies in current benchmarks, particularly in addressing multimodal interactions like visual misinterpretations and reasoning errors tied to diagram-based tasks.
    \item Validation of the necessity for distinct categories to capture errors unique to multimodal math problems.
\end{itemize}

\subsubsection*{3. Collection of Feedback from Students and Teachers}
The education team collected qualitative and quantitative feedback from students and teachers to ensure that the proposed error categories were grounded in real-world educational needs. Focus groups, surveys, and interviews were used to gather perspectives on common error patterns encountered during classroom activities and assessments.

\textbf{Key Insights:}
\begin{itemize}
    \item Teachers highlighted frequent calculation errors (\textbf{CAL}) and reasoning errors (\textbf{REAS}) as significant roadblocks to effective problem-solving.
    \item Students often reported confusion stemming from visual misinterpretations (\textbf{VIS}) and misunderstanding the question intent (\textbf{MIS}).
    \item Feedback emphasized the importance of separating reasoning-based errors from knowledge-based errors (\textbf{KNOW}) for better diagnostic support.
\end{itemize}

\subsubsection*{4. Second Round of Discussion and Alignment}
Following the feedback collection, the research and education teams reconvened to refine and align the error taxonomy. This phase involved iterative discussions to ensure that each category was distinct, comprehensive, and intuitive for annotators and end-users.

\textbf{Adjustments Made:}
\begin{itemize}
    \item Clarified the scope of \textbf{Reasoning Errors (REAS)} to focus on improper logical application rather than factual knowledge gaps.
    \item Strengthened the definition of \textbf{Visual Perception Errors (VIS)} to address multimodal-specific challenges, such as interpreting diagrams or image-based data.
    \item Enhanced examples for each category to support annotation clarity.
\end{itemize}

\subsubsection*{5. Initial Finalization and Feedback from Educational Organization}
The refined error categories were presented to a partner educational organization for feedback. This organization, which specializes in global education assessments, conducted an independent review and provided expert input.

\textbf{Key Outcomes:}
\begin{itemize}
    \item Positive validation of the categories' relevance and comprehensiveness.
    \item Minor recommendations, such as specifying units and signs in the \textbf{Calculation Errors (CAL)} category, were integrated.
\end{itemize}

\subsection*{6. Final Validation and Alignment with Annotation Team}
After incorporating feedback, the final set of error categories was finalized. The annotation team, comprising educational experts, received detailed guidelines and training to ensure consistent application of the taxonomy during the annotation process. Mock annotations were conducted to test the clarity and usability of the categories.

\textbf{Final Adjustments:}
\begin{itemize}
    \item Annotators highlighted the need for clearer boundaries between \textbf{Reasoning Errors (REAS)} and \textbf{Knowledge Errors (KNOW)}, leading to additional examples and decision rules in the annotation guidelines.
    \item Alignment meetings ensured that all discrepancies and ambiguities were resolved before the dataset’s official annotation began.
\end{itemize}

The aforementioned development process ensured that the five categories are comprehensive, robust, and applicable to both multimodal tasks and real-world educational scenarios.

\section{Additional Experimental Details}
\label{app:experiment}

\subsection{More Main Results}
\label{app:main_result}

\begin{figure*}[h!]
  \centering
  \includegraphics[width=0.8\textwidth]{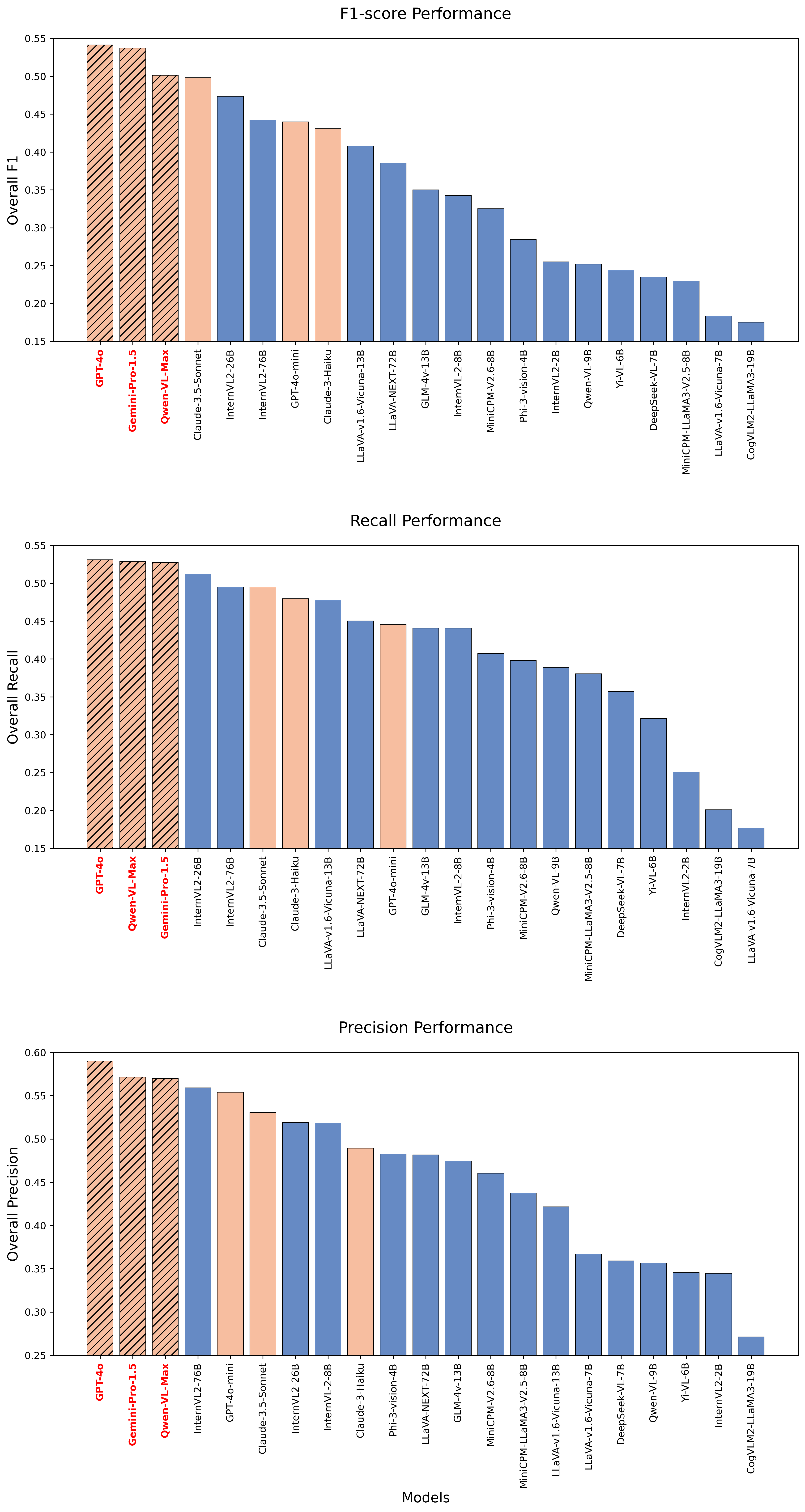}
  \caption{Error category performance of all models for F1, recall, and precision, respectively.}
\label{fig:cate_performance_bar_all}
\end{figure*}

\begin{figure*}[h!]
  \centering
  \includegraphics[width=0.8\textwidth]{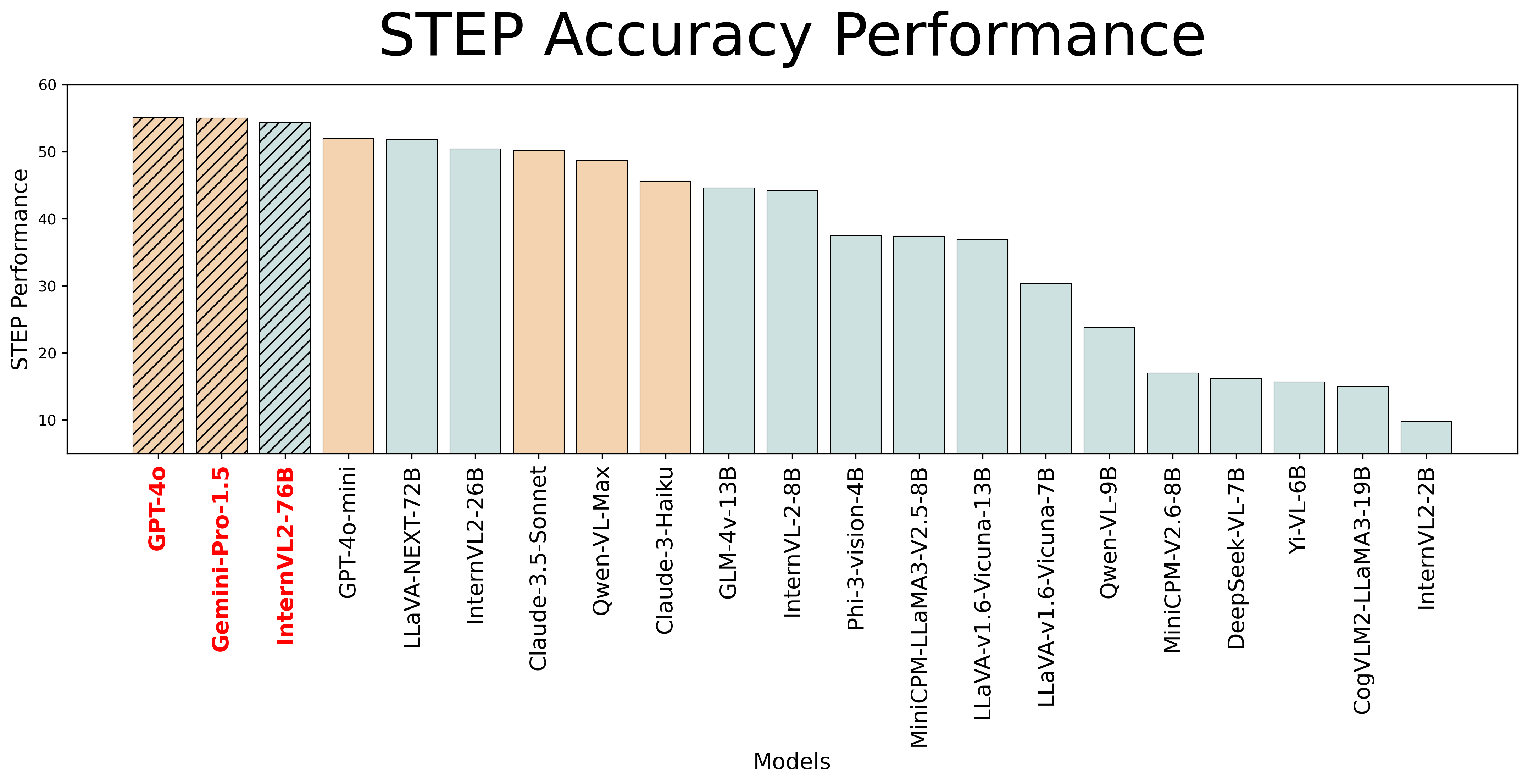}
  \caption{Error step performance of all models.}
\label{fig:step_performance_bar_all}
\end{figure*}

In this section, Figure \ref{fig:cate_performance_bar_all} shows the \textbf{overall error category} performance of all models for F1, recall, and precision, respectively. Figure \ref{fig:step_performance_bar_all} shows the \textbf{overall error step} performance of all models. Tables \ref{tab:recall_summary}, \ref{tab:prec_summary}, and \ref{tab:f1_summary} show the \textbf{category-level error category} performance for recall, precision, and F1, respectively.

\begin{table*}[h]
\centering
\small
\renewcommand\tabcolsep{2.5pt} 
\renewcommand\arraystretch{0.95} 
\resizebox{0.6\linewidth}{!}{
\begin{tabular}{c|c|ccccc}
\toprule
\textbf{Multimodal Large Language Models} & \textbf{Parameters} & \header{\textbf{VIS}} & \header{\textbf{CAL}} & \header{\textbf{REAS}} & \header{\textbf{KNOW}} & \header{\textbf{MIS}} \\

\midrule
\multicolumn{7}{l}{\hfill \textit{Open-Source MLLMs}} \\
\midrule
InternVL2~\citep{chen2023internvl} & 2B    & 0.32 & 0.38 & 0.12 & 0.00 & 0.24 \\
Phi-3-vision~\citep{abdin2024phi3} & 4B    & 0.09 & \high{0.99} & 0.06 & 0.03 & 0.04 \\
Yi-VL~\citep{young2024yi}          & 6B    & 0.09 & 0.77 & 0.04 & 0.14 & 0.00 \\
DeepSeek-VL~\citep{lu2024deepseek} & 7B    & 0.04 & 0.90 & 0.00 & \high{0.28} & 0.06 \\
LLaVA-v1.6-Vicuna~\citep{liu2024llavanext} & 7B    & 0.40 & 0.14 & 0.08 & 0.00 & \best{0.55} \\
InternVL-2~\citep{chen2023internvl} & 8B    & 0.12 & \high{0.99} & 0.13 & 0.10 & 0.02 \\
MiniCPM-LLaMA3-V2.5~\citep{yao2024minicpm} & 8B    & 0.04 & \best{1.00} & 0.02 & 0.02 & 0.00 \\
MiniCPM-V2.6~\citep{yao2024minicpm} & 8B    & 0.11 & 0.87 & 0.12 & 0.10 & 0.17 \\
Qwen-VL~\citep{Qwen-VL}           & 9B    & 0.08 & \high{0.99} & 0.03 & 0.00 & 0.00 \\
GLM-4v~\citep{glm2024chatglm}     & 13B   & 0.02 & 0.92 & 0.25 & 0.00 & 0.00 \\
LLaVA-v1.6-Vicuna~\citep{liu2024llavanext} & 13B   & 0.00 & 0.74 & 0.53 & 0.00 & 0.02 \\
CogVLM2-LLaMA3~\citep{wang2023cogvlm} & 19B   & \high{0.43} & 0.33 & 0.00 & 0.13 & 0.00 \\
InternVL2~\citep{chen2023internvl} & 26B   & 0.39 & 0.84 & 0.35 & 0.00 & 0.10 \\
LLaVA-NEXT~\citep{liu2024llavanext} & 72B   & 0.07 & 0.85 & 0.31 & 0.07 & 0.00 \\
InternVL2~\citep{chen2023internvl} & 76B   & 0.33 & 0.92 & 0.25 & 0.10 & 0.08 \\

\midrule
\multicolumn{7}{l}{\hfill \textit{Closed-Source MLLMs}} \\
\midrule
Qwen-VL-Max~\citep{Qwen-VL}       & -     & 0.15 & 0.78 & 0.50 & 0.14 & 0.36 \\
Claude-3-Haiku~\citep{claude3}    & -     & 0.10 & 0.77 & 0.46 & 0.04 & 0.01 \\
Claude-3.5-Sonnet~\citep{claude35} & -    & 0.35 & 0.48 & \best{0.64} & 0.21 & 0.11 \\
Gemini-Pro-1.5~\citep{reid2024gemini} & -  & \high{0.43} & 0.55 & \high{0.63} & 0.18 & 0.13 \\
GPT-4o-mini~\citep{openai2024gpt4omini} & - & 0.09 & 0.46 & 0.62 & \best{0.31} & 0.13 \\
GPT-4o~\citep{openai2024gpt4o}    & -     & \best{0.46} & 0.50 & \best{0.64} & 0.09 & \high{0.46} \\

\midrule
\multicolumn{7}{l}{\hfill \textit{Human Performance}} \\
\midrule
Human       & -     & 0.67 & 0.76 & 0.48 & 0.35 & 0.54 \\

\bottomrule
\end{tabular}
}
\caption{Comparison of open-source and closed-source MLLM performance (\textbf{recall} in percentage) across error detection tasks. We also denote \textbf{VIS}, \textbf{CAL}, \textbf{REAS}, \textbf{KNOW}, and \textbf{MIS} for visual perception error, calculation error, reasoning error, knowledge error, and misinterpretation of the question. The highest and second highest scores among MLLMs in each column are highlighted in \colorbox{red!20}{red} and \colorbox{blue!20}{blue}, respectively.}
\label{tab:recall_summary}
\end{table*}

\begin{table*}[h]
\centering
\small
\renewcommand\tabcolsep{2.5pt} 
\renewcommand\arraystretch{0.95} 
\resizebox{0.6\linewidth}{!}{
\begin{tabular}{c|c|ccccc}
\toprule
\textbf{Multimodal Large Language Models} & \textbf{Parameters} & \header{\textbf{VIS}} & \header{\textbf{CAL}} & \header{\textbf{REAS}} & \header{\textbf{KNOW}} & \header{\textbf{MIS}} \\

\midrule
\multicolumn{7}{l}{\hfill \textit{Open-Source MLLMs}} \\
\midrule
InternVL2~\citep{chen2023internvl} & 2B    & 0.11 & 0.43 & 0.43 & 0.00 & 0.11 \\
Phi-3-vision~\citep{abdin2024phi3} & 4B    & 0.40 & 0.40 & \high{0.65} & 0.26 & 0.22 \\
Yi-VL~\citep{young2024yi}          & 6B    & 0.23 & 0.37 & 0.43 & 0.05 & 0.00 \\
DeepSeek-VL~\citep{lu2024deepseek} & 7B    & 0.16 & 0.41 & 0.44 & 0.10 & 0.11 \\
LLaVA-v1.6-Vicuna~\citep{liu2024llavanext} & 7B    & 0.21 & 0.42 & 0.45 & 0.00 & 0.05 \\
InternVL-2~\citep{chen2023internvl} & 8B    & 0.62 & 0.41 & 0.61 & 0.33 & 0.33 \\
MiniCPM-LLaMA3-V2.5~\citep{yao2024minicpm} & 8B    & 0.61 & 0.37 & 0.51 & 0.15 & 0.00 \\
MiniCPM-V2.6~\citep{yao2024minicpm} & 8B    & 0.54 & 0.42 & 0.55 & 0.08 & 0.13 \\
Qwen-VL~\citep{Qwen-VL}           & 9B    & 0.39 & 0.39 & 0.36 & 0.00 & 0.25 \\
GLM-4v~\citep{glm2024chatglm}     & 13B   & 0.76 & 0.41 & 0.52 & 0.00 & 0.00 \\
LLaVA-v1.6-Vicuna~\citep{liu2024llavanext} & 13B   & 0.00 & 0.48 & 0.47 & 0.00 & 0.37 \\
CogVLM2-LLaMA3~\citep{wang2023cogvlm} & 19B   & 0.13 & 0.39 & 0.26 & 0.04 & 0.00 \\
InternVL2~\citep{chen2023internvl} & 26B   & 0.53 & 0.48 & 0.57 & 0.33 & 0.44 \\
LLaVA-NEXT~\citep{liu2024llavanext} & 72B   & 0.66 & 0.42 & 0.51 & \high{0.45} & 0.08 \\
InternVL2~\citep{chen2023internvl} & 76B   & 0.55 & 0.45 & \best{0.66} & \best{0.56} & \best{0.52} \\

\midrule
\multicolumn{7}{l}{\hfill \textit{Closed-Source MLLMs}} \\
\midrule
Qwen-VL-Max~\citep{Qwen-VL}       & -     & \high{0.81} & 0.53 & 0.57 & 0.36 & 0.24 \\
Claude-3-Haiku~\citep{claude3}    & -     & 0.60 & 0.45 & 0.52 & 0.16 & 0.40 \\
Claude-3.5-Sonnet~\citep{claude35} & -    & 0.61 & \high{0.56} & 0.52 & 0.13 & \high{0.46} \\
Gemini-Pro-1.5~\citep{reid2024gemini} & -  & 0.73 & \high{0.56} & 0.55 & 0.41 & 0.37 \\
GPT-4o-mini~\citep{openai2024gpt4omini} & - & \best{0.90} & 0.54 & 0.51 & 0.08 & 0.27 \\
GPT-4o~\citep{openai2024gpt4o}    & -     & 0.80 & \best{0.61} & 0.55 & 0.44 & 0.17 \\

\midrule
\multicolumn{7}{l}{\hfill \textit{Human Performance}} \\
\midrule
Human       & -     & 0.64 & 0.68 & 0.64 & 0.23 & 0.38 \\

\bottomrule
\end{tabular}
}
\caption{Comparison of open-source and closed-source MLLM performance (\textbf{precision} in percentage) across error detection tasks. We also denote \textbf{VIS}, \textbf{CAL}, \textbf{REAS}, \textbf{KNOW}, and \textbf{MIS} for visual perception error, calculation error, reasoning error, knowledge error, and misinterpretation of the question. The highest and second highest score among MLLMs in each column are highlighted in \colorbox{red!20}{red} and \colorbox{blue!20}{blue}, respectively.}
\label{tab:prec_summary}
\end{table*}

\begin{table*}[h]
\centering
\small
\renewcommand\tabcolsep{2.5pt} 
\renewcommand\arraystretch{0.95} 
\resizebox{0.6\linewidth}{!}{
\begin{tabular}{c|c|ccccc}
\toprule
\textbf{Multimodal Large Language Models} & \textbf{Parameters} & \header{\textbf{VIS}} & \header{\textbf{CAL}} & \header{\textbf{REAS}} & \header{\textbf{KNOW}} & \header{\textbf{MIS}} \\

\midrule
\multicolumn{7}{l}{\hfill \textit{Open-Source MLLMs}} \\
\midrule
InternVL2~\citep{chen2023internvl} & 2B    & 0.16 & 0.40 & 0.19 & 0.00 & 0.15 \\
Phi-3-vision~\citep{abdin2024phi3} & 4B    & 0.15 & 0.57 & 0.12 & 0.05 & 0.06 \\
Yi-VL~\citep{young2024yi}          & 6B    & 0.13 & 0.50 & 0.08 & 0.08 & 0.00 \\
DeepSeek-VL~\citep{lu2024deepseek} & 7B    & 0.07 & 0.57 & 0.00 & 0.15 & 0.08 \\
LLaVA-v1.6-Vicuna~\citep{liu2024llavanext} & 7B    & 0.28 & 0.22 & 0.14 & 0.00 & 0.09 \\
InternVL-2~\citep{chen2023internvl} & 8B    & 0.20 & 0.59 & 0.22 & 0.16 & 0.04 \\
MiniCPM-LLaMA3-V2.5~\citep{yao2024minicpm} & 8B    & 0.07 & 0.54 & 0.04 & 0.04 & 0.00 \\
MiniCPM-V2.6~\citep{yao2024minicpm} & 8B    & 0.18 & 0.56 & 0.19 & 0.08 & 0.15 \\
Qwen-VL~\citep{Qwen-VL}           & 9B    & 0.14 & 0.56 & 0.06 & 0.00 & 0.01 \\
GLM-4v~\citep{glm2024chatglm}     & 13B   & 0.04 & 0.57 & 0.34 & 0.00 & 0.00 \\
LLaVA-v1.6-Vicuna~\citep{liu2024llavanext} & 13B   & 0.00 & 0.58 & 0.50 & 0.00 & 0.04 \\
CogVLM2-LLaMA3~\citep{wang2023cogvlm} & 19B   & 0.21 & 0.36 & 0.01 & 0.06 & 0.00 \\
InternVL2~\citep{chen2023internvl} & 26B   & 0.45 & \best{0.61} & 0.44 & 0.01 & 0.17 \\
LLaVA-NEXT~\citep{liu2024llavanext} & 72B   & 0.12 & 0.57 & 0.39 & 0.12 & 0.01 \\
InternVL2~\citep{chen2023internvl} & 76B   & 0.41 & \high{0.60} & 0.36 & \high{0.18} & 0.14 \\

\midrule
\multicolumn{7}{l}{\hfill \textit{Closed-Source MLLMs}} \\
\midrule
Qwen-VL-Max~\citep{Qwen-VL}       & -     & 0.25 & 0.20 & 0.53 & \best{0.25} & \best{0.29} \\
Claude-3-Haiku~\citep{claude3}    & -     & 0.17 & 0.57 & 0.49 & 0.06 & 0.03 \\
Claude-3.5-Sonnet~\citep{claude35} & -    & 0.45 & 0.51 & \high{0.58} & 0.16 & 0.18 \\
Gemini-Pro-1.5~\citep{reid2024gemini} & -  & \high{0.54} & 0.56 & \high{0.58} & \best{0.25} & 0.19 \\
GPT-4o-mini~\citep{openai2024gpt4omini} & - & 0.16 & 0.50 & 0.56 & 0.13 & 0.17 \\
GPT-4o~\citep{openai2024gpt4o}    & -     & \best{0.58} & 0.55 & \best{0.59} & 0.15 & \high{0.25} \\

\midrule
\multicolumn{7}{l}{\hfill \textit{Human Performance}} \\
\midrule
Human       & -     & 0.65 & 0.71 & 0.55 & 0.28 & 0.44 \\

\bottomrule
\end{tabular}
}
\caption{Comparison of open-source and closed-source MLLM performance (\textbf{F1} in percentage) across error detection tasks. We also denote \textbf{VIS}, \textbf{CAL}, \textbf{REAS}, \textbf{KNOW}, and \textbf{MIS} for visual perception error, calculation error, reasoning error, knowledge error, and misinterpretation of the question. The highest and second highest score among MLLMs in each column are highlighted in \colorbox{red!20}{red} and \colorbox{blue!20}{blue}, respectively.}
\label{tab:f1_summary}
\end{table*}

\subsection{Human Performance Evaluation}
\label{app:human}
In the Human Performance section, the evaluation involved three educational expert evaluators, each independently performing the two subtasks — error step identification and error categorization — on a set of multimodal math problems. To ensure the validity of their assessments, a rigorous cross-checking procedure was implemented. After the initial independent evaluations, the results from all three experts were compared for both the identification of error steps and the categorization of those errors. When discrepancies arose, particularly in cases where the experts disagreed on which step contained an error or how an error should be classified, a structured conflict resolution process was followed.

The cross-check process began with identifying areas of disagreement between the evaluators. These conflicts were discussed in a series of consensus meetings, where the evaluators would review the conflicting steps or categorizations in detail. Each expert provided their rationale, referencing the mathematical principles involved as well as the multimodal representations of the problems. Through open dialogue, the evaluators aimed to reach a consensus on the correct interpretation of the error.

In cases where consensus could not be easily achieved, a majority-vote system was employed. However, for particularly complex or ambiguous cases, an additional adjudicator — who did not participate in the initial evaluations but had equivalent expertise — was consulted to provide a final judgment. This adjudicator reviewed the contentious cases along with the evaluators' justifications, ensuring an unbiased final decision. The outcome of this process was the creation of a refined ground truth dataset that balanced expert knowledge with the goal of consistent and reliable error identification and categorization.

\subsection{Prompt for MLLM Evaluation}
\label{app:prompt}
You can refer to Figures \ref{fig:prompt step} and \ref{fig:prompt category} for prompt details.

\begin{figure*}[th!]
    \centering
    \begin{tcolorbox}[colback=white, colframe=black, enhanced jigsaw, listing only, listing options={basicstyle=\rmfamily}]
        \textbf{Task Definition:} You are an education expert proficient in K-12 mathematics. Your task is to identify the first step where the mistake occurred in the incorrect answer reasoning steps based on the following mathematical question (including the textual and visual parts), reference answer, and incorrect answer. \\[1em]
        \textbf{Output format:} \\
        Error Step: Step X \\[1em]
        \textbf{Below is the reference content you need to identify the error step:} \\
        Question Image: \{image\} \\
        Question Text: \{content\} \\
        Correct Answer: \{answer\} \\
        Incorrect Answer: \{user$\_$answer\} \\
        Incorrect Answer Reasoning Steps:\{user$\_$answer$\_$steps\} \\[1em]
        \textbf{Instruction:} Please provide the corresponding error step identification in the format "Error Step: Step X", without any additional content.
    \end{tcolorbox}
    \caption{Prompt for error step identification task.}
    \label{fig:prompt step}
\end{figure*}


\begin{figure*}[th!]
    \centering
    \begin{tcolorbox}[colback=white, colframe=black, enhanced jigsaw, listing only, listing options={basicstyle=\rmfamily}]
        \textbf{Task Definition:} You are an education expert proficient in K-12 mathematics. Your task is to identify the category of error for the incorrect answer based on the following question (including the textual and visual parts), reference answer, and incorrect answer. The error should belong to one of the following categories: Visual Perception Error, Reasoning Error, Knowledge Error, Calculation Error, or Misinterpretation of the Question. \\[1em]
        \textbf{Output format:} \\
        Error Category: Clearly indicate which error category it belongs to.\\[1em]
        \textbf{The definitions of the error categories are as follows:} \\
        \ding{79}Visual Perception Error: Failure to accurately obtain information from the images or charts in the question due to visual issues, leading to errors. \\
        \ding{79}Reasoning Error: Improper reasoning during the problem-solving process, failure to correctly apply logical relationships or draw conclusions, leading to errors \\
        \ding{79}Knowledge Error: Errors occur when applying relevant knowledge points due to incomplete or incorrect understanding of knowledge. \\
        \ding{79}Calculation Error: Errors occur in the calculation process, such as addition, subtraction, multiplication, division mistakes, or unit conversion errors, or errors in numerical symbols between multiple steps.\\
        \ding{79}Misinterpretation of the Question: Failure to correctly understand the requirements of the question or misinterpreting the meaning of the question stem, leading to an irrelevant answer, such as answering with numbers when letters are required, and vice versa. \\[1em]
        \textbf{Below is the reference content you need to identify the error step:} \\
        Question Image: \{image\} \\
        Question Text: \{content\} \\
        Correct Answer: \{answer\} \\
        Incorrect Answer: \{user$\_$answer\} \\
        Incorrect Answer Reasoning Steps:\{user$\_$answer$\_$steps\} \\[1em]
        \textbf{Instruction:} Please provide the corresponding error category in the format "Error Category: X", without any additional content.
    \end{tcolorbox}
    \caption{Prompt for error categorization task.}
    \label{fig:prompt category}
\end{figure*}


\subsection{Model Sources}
\label{app:sources}

Table \ref{tab:mllm hyperparams} details specific sources for the various MLLMs we evaluated. The hyperparameters for the experiments are set to their default values unless specified otherwise.

\begin{table*}[th!]
\small
\centering
\begin{tabular}{p{0.18\linewidth} | p{0.3\linewidth} | p{0.4\linewidth}}
\toprule
\textbf{MLLMs} & \textbf{Source} & \textbf{URL} \\
\midrule
InternVL2-2B & local checkpoint & \url{https://huggingface.co/OpenGVLab/InternVL2-2B} \\
\midrule
InternVL2-8B & local checkpoint & \url{https://huggingface.co/OpenGVLab/InternVL2-8B} \\
\midrule
InternVL2-26B & local checkpoint & \url{https://huggingface.co/OpenGVLab/InternVL2-26B} \\
\midrule
InternVL2-76B & local checkpoint &  \url{https://huggingface.co/OpenGVLab/InternVL2-Llama3-76B}\\
\midrule
Phi-3-vision-4B & local checkpoint & \url{https://huggingface.co/microsoft/Phi-3-vision-128k-instruct} \\
\midrule
Yi-VL-6B & local checkpoint & \url{https://huggingface.co/01-ai/Yi-VL-6B} \\
\midrule
DeepSeek-VL-7B & local checkpoint & \url{https://huggingface.co/deepseek-ai/deepseek-vl-7b-chat} \\
\midrule
LLaVA-v1.6-Vicuna-7B & local checkpoint & \url{https://huggingface.co/llava-hf/llava-v1.6-vicuna-7b-hf} \\
\midrule
LLaVA-v1.6-Vicuna-13B & local checkpoint & \url{https://huggingface.co/llava-hf/llava-v1.6-vicuna-13b-hf} \\
\midrule
LLaVA-NEXT-72B & local checkpoint & \url{https://huggingface.co/llava-hf/llava-next-72b-hf} \\
\midrule
MiniCPM-V2.5-8B & local checkpoint & \url{https://huggingface.co/openbmb/MiniCPM-Llama3-V-2_5} \\
\midrule
MiniCPM-V2.6-8B & local checkpoint & \url{https://huggingface.co/openbmb/MiniCPM-V-2_6} \\
\midrule
Qwen-VL-9B & local checkpoint & \url{https://huggingface.co/Qwen/Qwen-VL-Chat} \\
\midrule
GLM-4v-13B & local checkpoint & \url{https://huggingface.co/THUDM/glm-4v-9b} \\
\midrule
CogVLM2-19B & local checkpoint & \url{https://huggingface.co/THUDM/cogvlm2-llama3-chat-19B} \\
\midrule
Qwen-VL-Max & \texttt{qwen-vl-max-0809} & \url{https://modelscope.cn/studios/qwen/Qwen-VL-Max} \\
\midrule
Claude-3-Haiku & \texttt{claude-3-haiku} &  \url{https://www.anthropic.com/api}\\
\midrule
Claude-3.5-Sonnet & \texttt{claude-3-5-sonnet} &  \url{https://www.anthropic.com/api}\\
\midrule
Gemini-Pro-1.5 & \texttt{gemini-1.5-pro-latest} & \url{https://deepmind.google/technologies/gemini/pro/} \\
\midrule
GPT-4o-mini & \texttt{gpt-4o-mini-2024-07-18} & \url{https://platform.openai.com/docs/models/gpt-4o-mini} \\
\midrule
GPT-4o & \texttt{gpt-4o-2024-08-06} & \url{https://platform.openai.com/docs/models/gpt-4o}\\

\bottomrule
\end{tabular}
\caption{Sources of our evaluated MLLMs.}
\label{tab:mllm hyperparams}
\end{table*}

\subsection{Detailed Actionable Suggestions}
\label{app:actionable suggestions}
\textbf{Finding \#1: Closed-source MLLMs outperform open-source MLLMs} \\
\textbf{Actionable Suggestion:} In order to improve the performance of open-source MLLMs, it is crucial to focus on distilling the error detection capabilities of closed-source models \citep{hsieh2023distilling,liang2024module}. One effective approach is to use a \textit{teacher-student framework} in which the more powerful closed-source model acts as the teacher and the open-source model as the student. This method allows the open-source models to learn from the strengths of closed-source models, particularly in error detection tasks including error step identification and error categorization. Moreover, open-source models should leverage datasets used by closed-source models and augment their training process to better mimic the proprietary training regimens of these models \citep{liang2024module,aslam2024multi}. To optimize performance, fine-tuning should be guided by the insights from closed-source models, including how these models handle complex error categories and balance their performance across various error types. This approach will ensure a more robust open-source model, capable of addressing the current performance gaps. \\

\textbf{Finding \#2: Open-source MLLMs over-predict CAL category} \\
\textbf{Actionable Suggestion:} The tendency for open-source MLLMs to over-predict the CAL (Calculation Error) category is an issue that arises from their bias towards easier tasks. To address this, models should be regularized during training to reduce their preference for simpler categories like CAL. This can be achieved through the use of \textit{weighted loss functions}, such as \textit{Focal Loss} \citep{li2022generalized} or \textit{AdaFocal} \citep{ghosh2022adafocal}, which down-weight easier categories and force the model to focus on more challenging ones. Additionally, introducing \textit{class balancing techniques} such as oversampling the underrepresented categories (e.g., REAS, KNOW) or undersampling the CAL category can further help in addressing this bias \citep{ghosh2024class}. Another key approach is data augmentation, where the variety and complexity of error cases are increased, particularly for the more challenging categories. This will ensure that the model learns to classify all types of errors more evenly, avoiding an over-reliance on CAL \citep{iqbal2024data}. Finally, \textit{meta-learning} techniques can be employed to dynamically adjust the model’s bias toward different categories during training, allowing the model to better adapt to different error types \citep{vettoruzzo2024advances}. \\

\textbf{Finding \#3: STEP tasks are easier than CATE tasks} \\
\textbf{Actionable Suggestion:} The disparity in performance between STEP and CATE tasks suggests that MLLMs need to be trained to better handle the complexity of error categorization. Since STEP tasks primarily involve localizing specific errors, which is conceptually simpler than categorizing them, it is crucial to build a stronger relationship between the two tasks in the training process. One effective method is to use \textit{multi-task learning}, where the model is simultaneously trained on both STEP and CATE tasks, allowing it to learn not only how to localize errors but also how to classify them accurately \citep{chen2024multi,xin2024mmap}. Additionally, \textit{contrastive learning} can be used to distinguish between similar error steps and categories, improving the model’s ability to reason about the relationship between error localization and categorization. Training data should also be designed to emphasize this relationship, ensuring that the model can learn the necessary contextual understanding to categorize errors correctly. By focusing on these aspects, MLLMs will be better equipped to handle the more complex task of error categorization \citep{hu2024comprehensive}. \\

{\textbf{Finding \#4: CAL is the easiest category, while KNOW is the hardest} \\
\textbf{Actionable Suggestion:} The difficulty gap between CAL and KNOW errors highlights the need for specialized strategies to handle knowledge errors. Since CAL errors are generally more deterministic and easy to identify, while KNOW errors require deeper understanding and reasoning, MLLMs should incorporate domain-specific knowledge to improve their performance in this category. One approach is to integrate external knowledge bases or knowledge graphs into the training dataset, providing the model with richer, contextually relevant information. This will help the model recognize errors related to factual inaccuracies or incomplete reasoning \citep{sun2023think,pan2024unifying}. Additionally, knowledge-intensive reasoning models can be introduced to simulate more advanced human-like problem-solving capabilities. Techniques such as external validation of logical consistency can also be applied to better identify and rectify knowledge errors. Furthermore, few-shot learning methods can be used to allow MLLMs to generalize from limited examples, especially for rare or complex knowledge errors that are more difficult to detect \citep{ma2023fairness,ma2023large}. By improving the model’s access to domain-specific knowledge and reasoning tools, its ability to handle KNOW errors will be significantly enhanced. \\

\textbf{Finding \#5: Gap to human-level performance in error detection} \\
\textbf{Actionable Suggestion:} To bridge the gap between human-level performance and MLLM performance, particularly in tasks such as Visual Perception Errors (VIS) and Reasoning Errors (REAS), MLLMs need to be trained to better mimic human cognitive processes. One promising approach is to employ \textit{Reinforcement Learning from Human Feedback (RLHF)}, where human evaluators guide the model by providing corrective feedback and insights into error causes \citep{wang2024secrets,kirk2023understanding}. This will help the model align more closely with human reasoning mechanisms, particularly in tasks that require higher-level cognitive functions. In addition, models can be trained to simulate human visual perception by integrating attention mechanisms and vision-language models that enable more sophisticated visual error detection. Incorporating logical reasoning modules into the model will also improve its performance in REAS, allowing it to understand the logical flow of the problem and detect reasoning errors more effectively. Finally, cross-modal alignment between text and image modalities will ensure that MLLMs process visual and textual inputs in a more integrated and human-like manner, thereby improving performance in VIS error detection \citep{shen2023cross}. By aligning MLLMs more closely with human cognitive processes, it is possible to achieve significant improvements in error detection tasks and approach human-level performance. \\

\textbf{Finding \#6: Best Generalist models outperform specialized ones} \\
\textbf{Actionable Suggestion:} To enhance the performance of specialized reasoning and math models on complex multimodal error analysis tasks, developers should prioritize strategies that broaden their contextual understanding and error analysis capabilities beyond their narrow domain. One key action is to augment their training datasets with a more diverse range of multimodal examples that explicitly feature varied error types and require nuanced, cross-modal reasoning for identification \cite{li2023query,yin2024mumath,you2024mumath}. This could involve curating or synthetically generating data that forces models to not just solve a problem, but to also analyze and explain potential errors in presented solutions, mirroring the capabilities of generalist models. Furthermore, incorporating instruction tuning with fine-grained error analysis prompts, similar to how general-purpose visual language models are trained \cite{liu2023visual}, can help specialized models develop a more robust understanding of error patterns. Finally, exploring hybrid architectures or ensemble methods, where a specialized model’s deep domain knowledge is guided or supplemented by a generalist model’s broader contextual awareness and error-spotting acumen, could offer a practical path to improved performance without sacrificing specialization entirely \cite{xu2025towards,bi2025reasoning,yan2025mathagent}.

\subsection{CAL and non-CAL Distribution of MLLMs}
\label{app:cal_distribution}
In this section, we indicate the distribution of CAL and non-CAL category predictions of 21 representative MLLMs, as shown in Figure \ref{fig:cal_appendix}. It can be seen that there is a bias towards CAL category among most open-source MLLMs, while closed-source ones except for Claude-3-Haiku and Qwen-VL-Max do not have such a bias for error categorization task.
\begin{figure*}[th!]
  \centering
  \includegraphics[width=\textwidth]{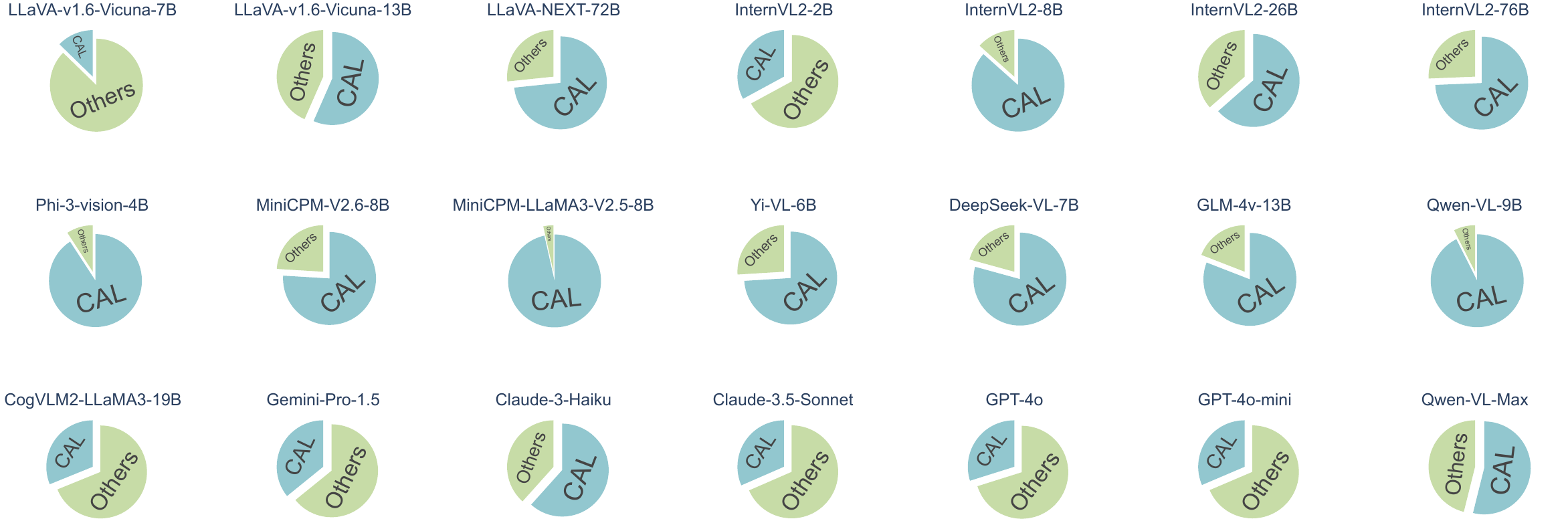}
  \vspace{-3mm}
  \caption{Distribution of CAL and non-CAL category predictions of all MLLMs we evaluate.}
\label{fig:cal_appendix}
\end{figure*}

\subsection{Visual Perception Analysis}
\label{app:visual_analysis}

\begin{figure}[ht]
    \centering
    \includegraphics[width=1\linewidth]{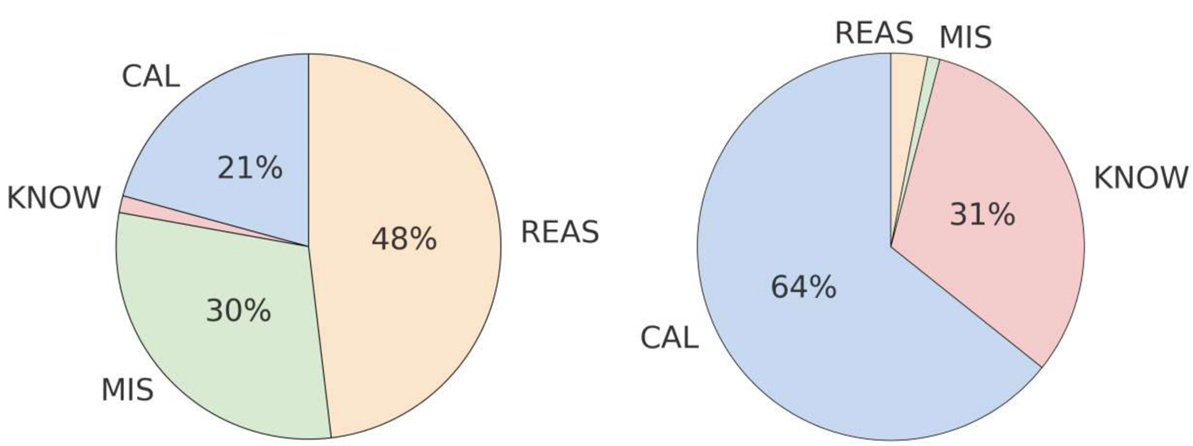}
    \caption{The error category distribution of misjudged VIS cases of GPT-4o (left) and CogVLM2-LLaMA3 (right).}
    \label{fig:vis bad case distribution pie}
    \vspace{-4mm}
\end{figure}

\textbf{Finding \#1: Closed-source MLLMs are most likely to misjudge VIS as REAS in error categorization task.} Taking GPT-4o model as an example, as shown in the Figure \ref{fig:vis bad case distribution pie}, 48\% of VIS are misclassified as REAS, followed by 30\% being misjudged as MIS. When MLLM needs to handle information involving both visual and linguistic elements simultaneously, if an erroneous response to a math query originates from VIS, it mistakenly attributes this to a flaw in logical reasoning that occurs subsequent to initial visual misinterpretation.

\textbf{Finding \#2: Open-source MLLMs are more likely to misclassify VIS as CAL.} Taking the open-source model CogVLM2-LLaMA3, which performs best in identifying VIS, as an example, CAL accounts for 64\% of misclassified category, as illustrated in the Figure \ref{fig:vis bad case distribution pie}. When handling complex visual information, especially in geometry problems, the MLLM often struggles to accurately extract key features. Due to the open-source MLLM's weaker multimodal integration capabilities, it simplifies visual issues into numerical calculation problems. The lack of sufficient training and data for visual-related errors is also a key reason behind this phenomenon \citep{wichmann2023deep}. See more analysis on misclassification for each category and visual perception case study in Appendix \ref{app:confusion_matrix} and \ref{app:vis_bad_category_gpt}.

\subsection{Analysis of Confusion Matrix for CATE task}
\label{app:confusion_matrix}
Figures \ref{fig:confusion matrix internvl} and \ref{fig:confusion matrix gpt4o} present the confusion matrices for InternVL2-76B and GPT-4o, two MLLMs evaluated on five error categories. The matrices show the count of predictions for each category, with diagonal entries representing correct predictions and off-diagonal entries indicating misclassifications. These visualizations provide insights into each model's strengths and weaknesses.

InternVL2-76B shows strong performance in detecting CAL, with 843 correct predictions, indicating its robust numerical reasoning capability. However, the model struggles to distinguish between REAS and CAL, misclassifying 626 REAS instances as CAL. This confusion suggests an over-reliance on numerical features and an inability to separate logical reasoning tasks from computational ones. Additionally, there is significant misclassification of VIS into CAL, with 244 cases, highlighting a potential weakness in integrating visual and textual modalities. These trends may stem from InternVL2-76B’s limited domain-specific reasoning ability.

GPT-4o, on the other hand, demonstrates relatively good performance in VIS, with 183 correct predictions, significantly outperforming InternVL2-76B. Its capability in REAS is also notable, with 617 correct predictions, suggesting a more balanced reasoning ability. However, GPT-4o struggles more with CAL, achieving only 460 correct predictions, and shows significant confusion between CAL and REAS, with 299 CAL instances misclassified as REAS. Furthermore, the model has difficulty with MIS, misclassifying 45 MIS cases as REAS, pointing to challenges in identifying nuanced interpretational issues. These trends suggest that GPT-4o’s emphasis on multimodal alignment and contextual understanding contributes to its strengths in VIS and REAS but comes at the expense of CAL performance.

Comparing the two models reveals distinct strengths and weaknesses. GPT-4o significantly outperforms InternVL2-76B in VIS, likely due to superior multimodal visual-text alignment capabilities. Both models exhibit confusion between REAS and CAL, but GPT-4o shows a more balanced classification ability in REAS. MIS remains a challenging category for both models, though GPT-4o struggles slightly more in distinguishing it from REAS. These differences may arise from variations in model architecture and training objectives. This analysis underscores the complementary strengths of these models: InternVL2-76B excels in numerical reasoning, while GPT-4o performs better in visual perception and logical reasoning. Future research could explore ways to integrate their strengths for a more robust multimodal error detection system.

\begin{figure*}[th!]
    \centering
    \begin{minipage}[t]{0.45\textwidth}
     \centering
     \includegraphics[width=\linewidth]{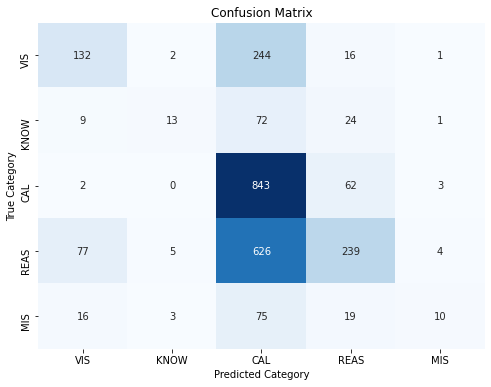}
      \caption{The confusion matrix of five error categories predicted by InternVL2-76B, the open-source MLLM with the best overall performance on error detection.}
      \label{fig:confusion matrix internvl}
    \end{minipage}
    \hfill
    \begin{minipage}[t]{0.45\textwidth}
     \centering
     \includegraphics[width=\linewidth]{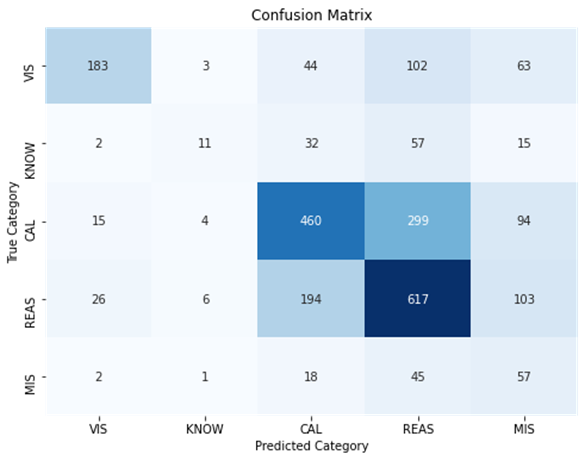}
      \caption{The confusion matrix of five error categories predicted by GPT-4o, the closed-source MLLM with the best overall performance on error detection.}
      \label{fig:confusion matrix gpt4o}
    \end{minipage}
\end{figure*}



\subsection{Visual Bad Cases Predicted by GPT-4o}
\label{app:vis_bad_category_gpt}

Visual perception errors are critical in multimodal error detection tasks, as they impact the accurate comprehension of mathematical problems presented with both text and diagrams. As illustrated in Figures \ref{fig:gpt visual bad case category distance}, \ref{fig:gpt visual bad case category diagram}, \ref{fig:gpt visual bad case category spatial}, \ref{fig:gpt visual bad case category flip} and \ref{fig:gpt visual bad case category shape}, the five primary categories of visual errors observed in GPT-4o (the MLLM with best overall and VIS performance) include \textbf{distance perception}, \textbf{diagram perception}, \textbf{spatial perception}, \textbf{flip/fold perception}, and \textbf{shape perception}. These categories differ in their cognitive demands: distance perception focuses on point identification; diagram perception on quantitative estimation; spatial perception on geometric visualization; flip/fold perception on mental rotation; and shape perception on object classification \citep{lu2023mathvista,zhang2024mathverse}. Detecting such errors is challenging because they often require both intricate visual processing and precise interpretation of mathematical relations, which can be difficult to encode in current MLLMs. To overcome these challenges, future MLLMs should incorporate more advanced visual reasoning capabilities, possibly through enhanced alignment between vision and language modalities, enabling better detection and correction of complex perception errors \citep{song2023bridge}. This could significantly improve the robustness of MLLMs in mathematical and other perception-heavy tasks.

\begin{figure*}[h!]
  \centering
  \includegraphics[width=0.8\textwidth]{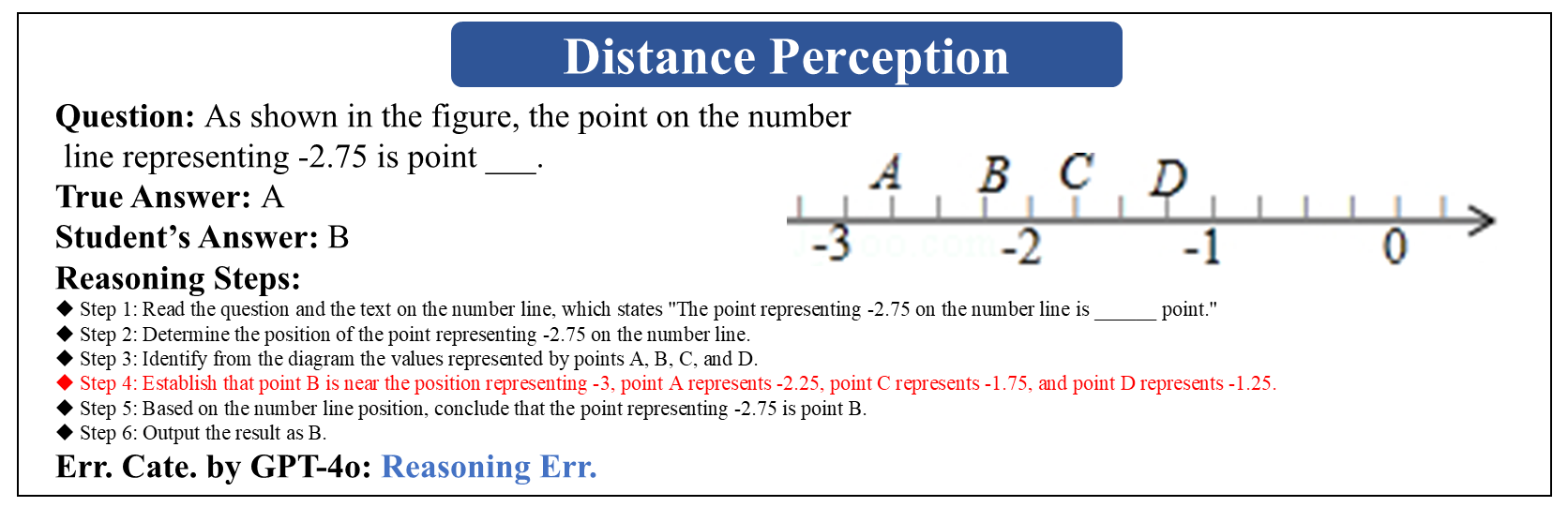}
  \caption{Distance bad case where GPT-4o predicts visual perception errors incorrectly.}
\label{fig:gpt visual bad case category distance}
\end{figure*}

\begin{figure*}[h!]
  \centering
  \includegraphics[width=0.8\textwidth]{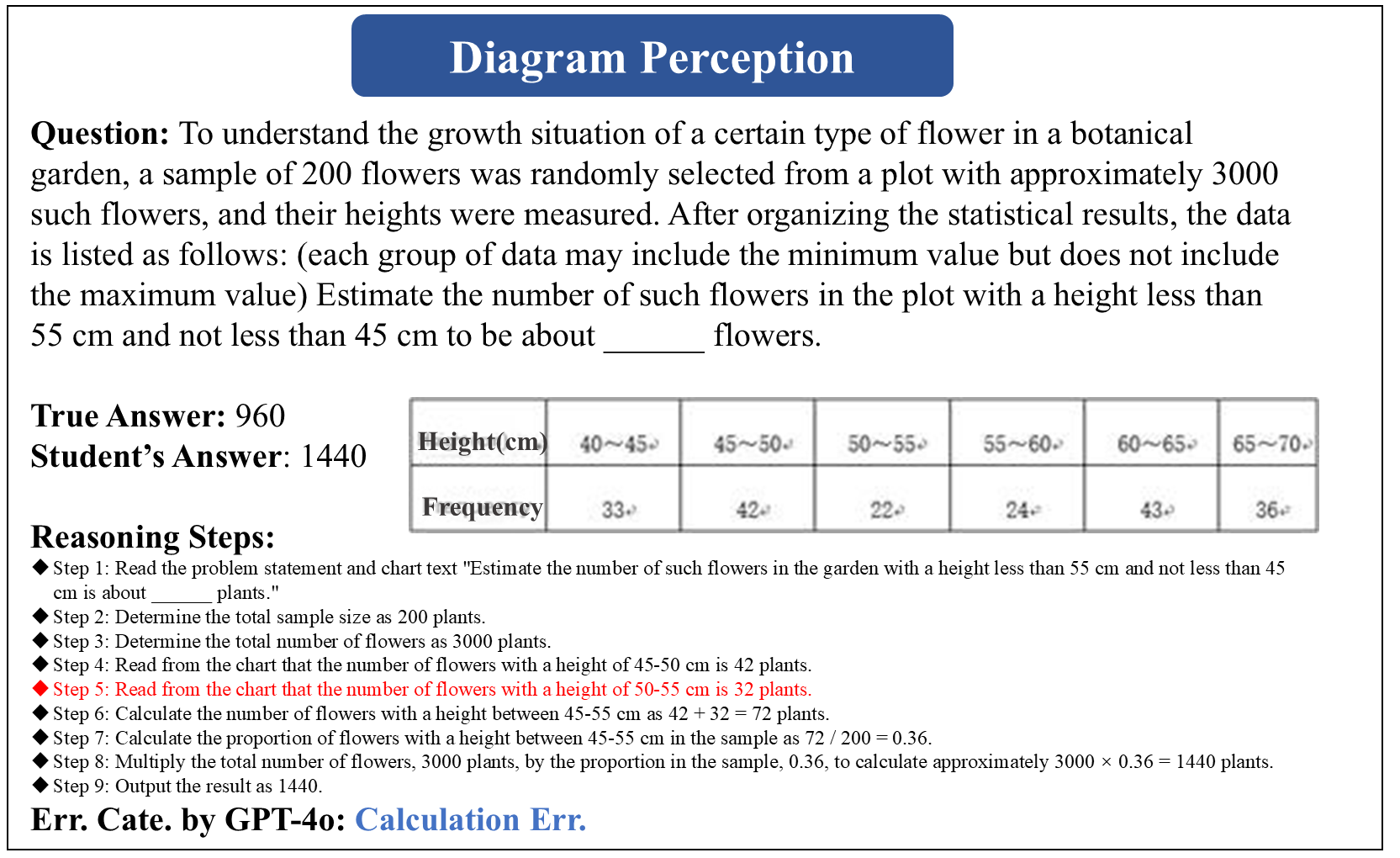}
  \caption{Diagram bad case where GPT-4o predicts visual perception errors incorrectly.}
\label{fig:gpt visual bad case category diagram}
\end{figure*}

\begin{figure*}[h!]
  \centering
  \includegraphics[width=0.8\textwidth]{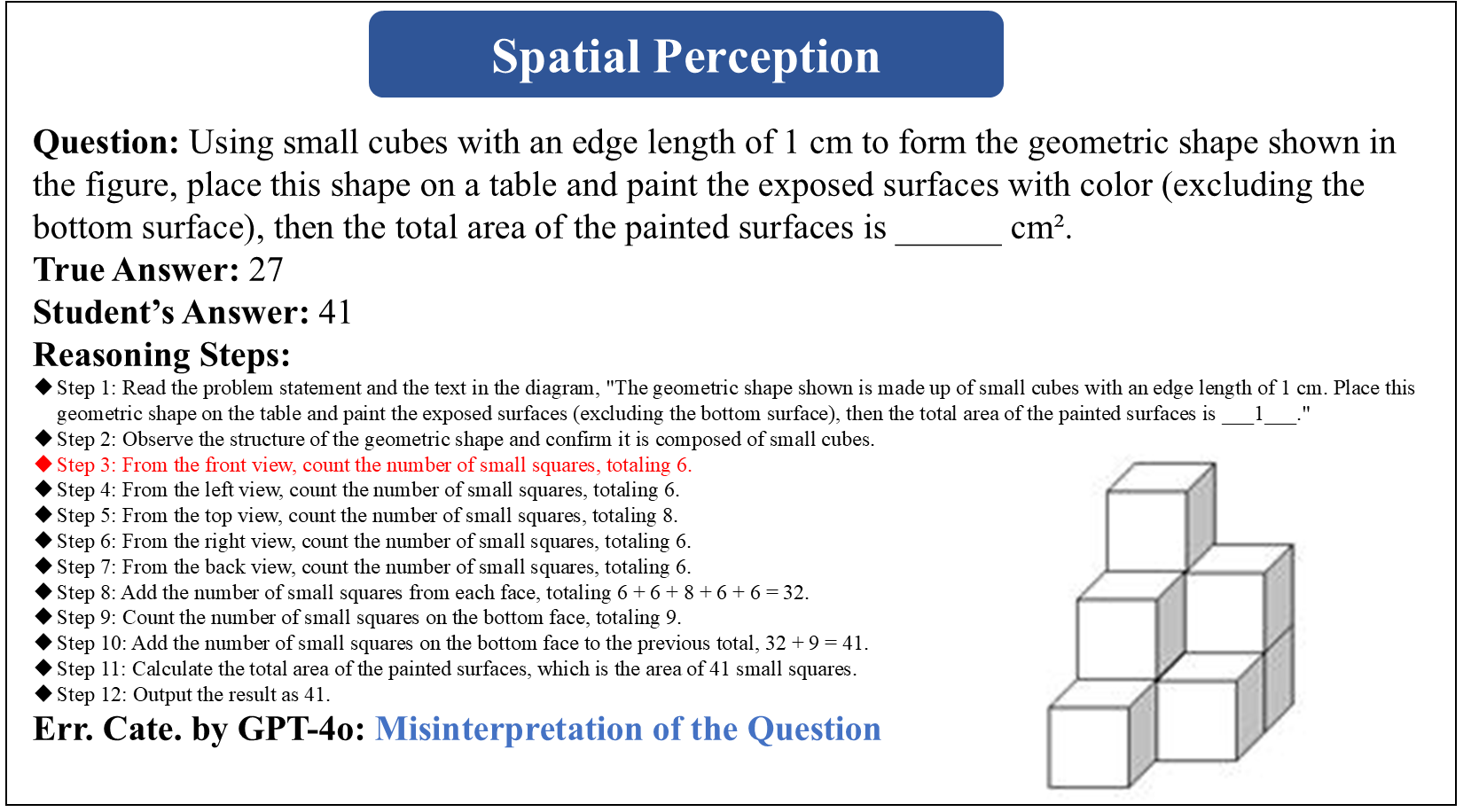}
  \caption{Spatial bad case where GPT-4o predicts visual perception errors incorrectly.}
\label{fig:gpt visual bad case category spatial}
\end{figure*}

\begin{figure*}[h!]
  \centering
  \includegraphics[width=0.8\textwidth]{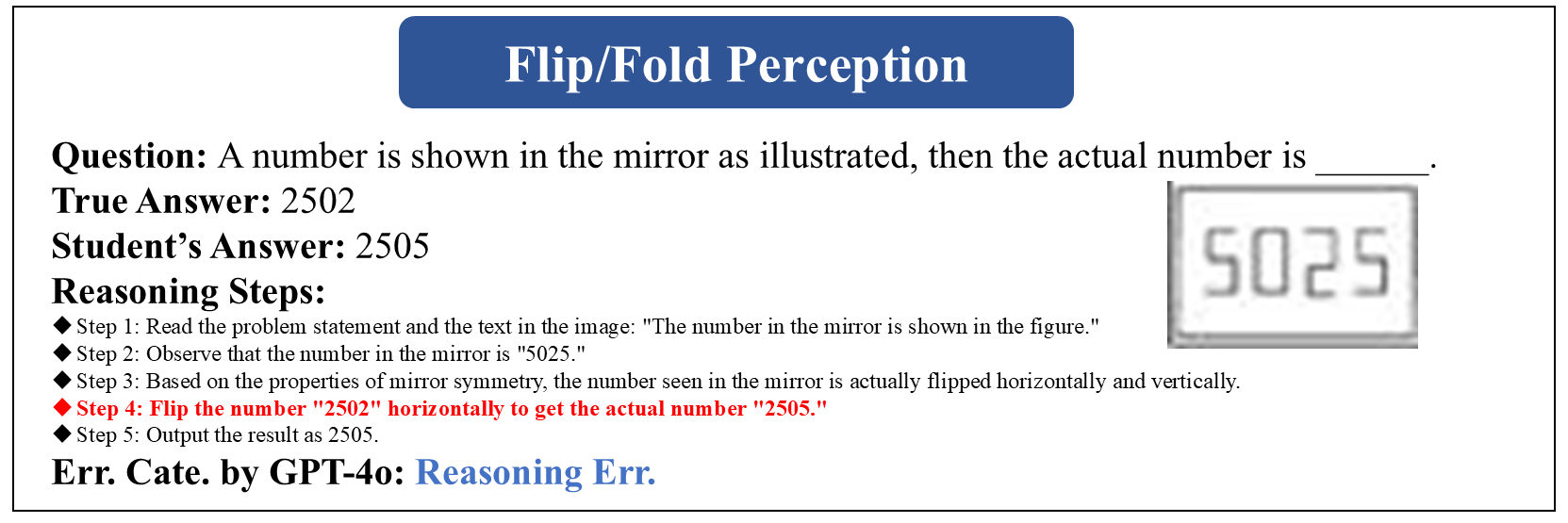}
  \caption{Flip \& fold bad case where GPT-4o predicts visual perception errors incorrectly.}
\label{fig:gpt visual bad case category flip}
\end{figure*}

\begin{figure*}[h!]
  \centering
  \includegraphics[width=0.8\textwidth]{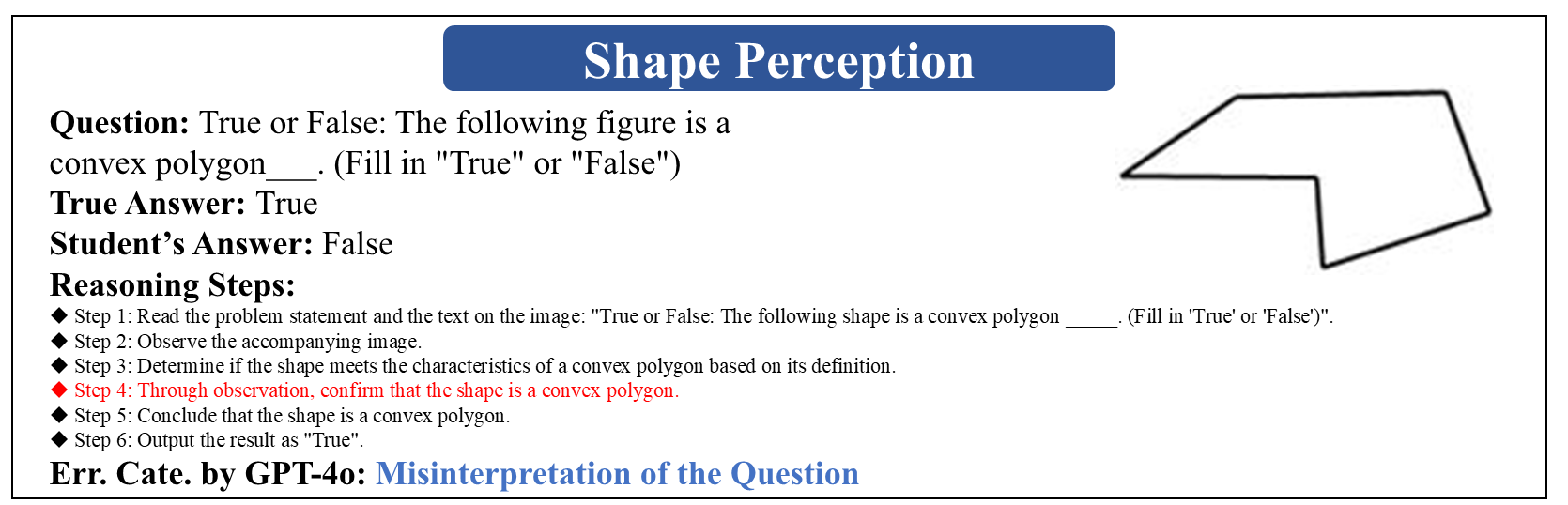}
  \caption{Shape bad case where GPT-4o predicts visual perception errors incorrectly.}
\label{fig:gpt visual bad case category shape}
\end{figure*}

\subsection{Relation between Error Category and Error Step}
\label{app:step_percentage}

\begin{figure}[!t]
    \centering
    \includegraphics[width=0.8\linewidth]{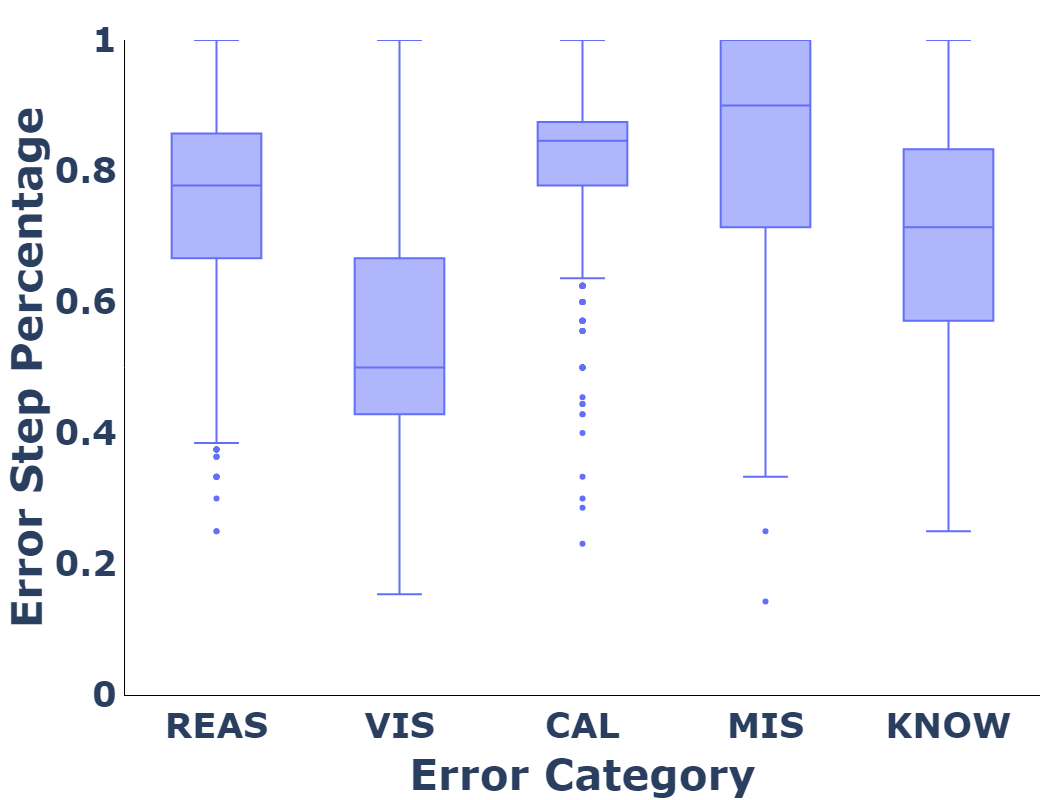}
    \caption{The error step distribution (in percentage) of error categories in \dataset dataset.}
    \label{fig:step percentage}
\end{figure}

\textbf{Finding \#1: There is a close relationship between different error category and their distribution in the reasoning steps.} As shown in Figure \ref{fig:step percentage}, VIS tends to occur in the earlier to mid-stages, accounting for a median proportion of 0.5 of total steps. In contrast, MIS, REAS, CAL, and KNOW are more likely to arise in the later stages, with their median proportions ranging from 0.7 to 0.9. More analysis of this relationship across MLLMs in terms of cognitive load analysis can be seen in Appendix \ref{app:cognitive_load}.

\textbf{Finding \#2: VIS occurs in the earlier stages of problem-solving reasoning}. This finding could be closely linked to the sequence in which students approach the task \citep{binz2023turning,kennedy2024cognitive}. Since image content often serves as a key reference early on, any misinterpretation of this visual information directly impacts the subsequent problem-solving steps. Students typically first examine the image, and then integrate the information before proceeding to reasoning or calculation. As a result, visual perception errors arise earlier compared to other types of errors.


\textbf{Finding \#3: Other error categories are primarily in later stages of problem-solving reasoning}. This may be linked to the increasing cognitive load students encounter during problem-solving. Cognitive Load Theory posits that information complexity ranges from low to high interactivity \citep{cognitiveloadtheory,binz2023turning}. While low-interactivity information can be understood independently, high-interactivity information requires simultaneous processing of related elements, thus increasing cognitive load \citep{kennedy2024cognitive,abbad2023complexity}. In later stages, students must integrate complex information from multiple sources. For instance, calculating the distance between two points needs increasing interactivity heightens cognitive load, leading to errors like forgetting to take the square root or miscalculating differences. Consequently, as cognitive load rises, the frequency of errors in later steps also increases.

\subsection{Cognitive Load Analysis Across MLLMs}
\label{app:cognitive_load}
In analyzing the error step distribution for the multimodal error detection task using InternVL2-76B (see Figure \ref{fig:step percentage internvl}) and GPT-4o (see Figure \ref{fig:step percentage gpt4o}), we observe a consistency in the pattern of error category distribution across both MLLM's predictions and those in \dataset (see Figure \ref{fig:step percentage}). In particular, VIS tends to occur in the earlier stages of problem-solving for both MLLMs, which aligns with the sequence in which students typically approach tasks. Since visual content often serves as a key reference at the outset, any misinterpretation of this information can significantly impact subsequent steps. Students generally examine the image first and then integrate the information before proceeding to reasoning or calculation, leading to visual perception errors arising earlier compared to other types of errors.

Other error categories, such as REAS, CAL, MIS, and KNOW, are more likely to emerge in the later stages of problem-solving. This pattern is linked to the increasing cognitive load students encounter as they progress. According to Cognitive Load Theory, information complexity ranges from low to high interactivity. Low-interactivity information can be understood independently, whereas high-interactivity information requires the simultaneous processing of related elements, thereby increasing cognitive load. In the later stages, students must integrate complex information from multiple sources, which can lead to errors like forgetting to take the square root or miscalculating differences when calculating distances, for example. Consequently, the frequency of errors in later steps increases with the rising cognitive load.

Despite the overall pattern being consistent, there may be subtle differences between InternVL2-76B and GPT-4o in terms of error step distribution, especially for MIS category. These differences could be attributed to the models' distinct architectures and training data, which might influence their approaches to error detection. As an open-source MLLM, InternVL2-76B might not have been optimized for specific types of questions or educational contexts, which could lead to a higher variability in MIS.

\begin{figure}[th!]
    \centering
    \begin{minipage}[t]{0.45\textwidth}
     \centering
     \includegraphics[width=\linewidth]{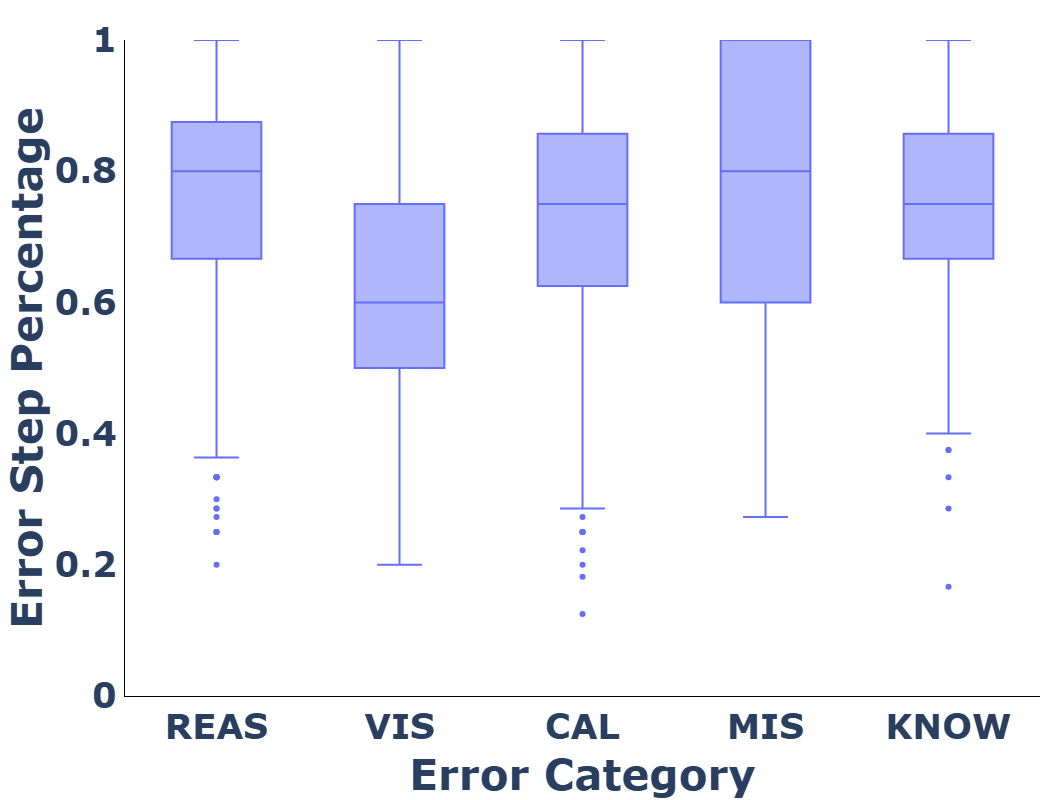}
      \caption{The error step distribution (in percentage) of error categories predicted by InternVL2-76B, the open-source MLLM with the best overall performance on error detection.}
      \label{fig:step percentage internvl}
    \end{minipage}
    \hfill
    \begin{minipage}[t]{0.45\textwidth}
     \centering
     \includegraphics[width=\linewidth]{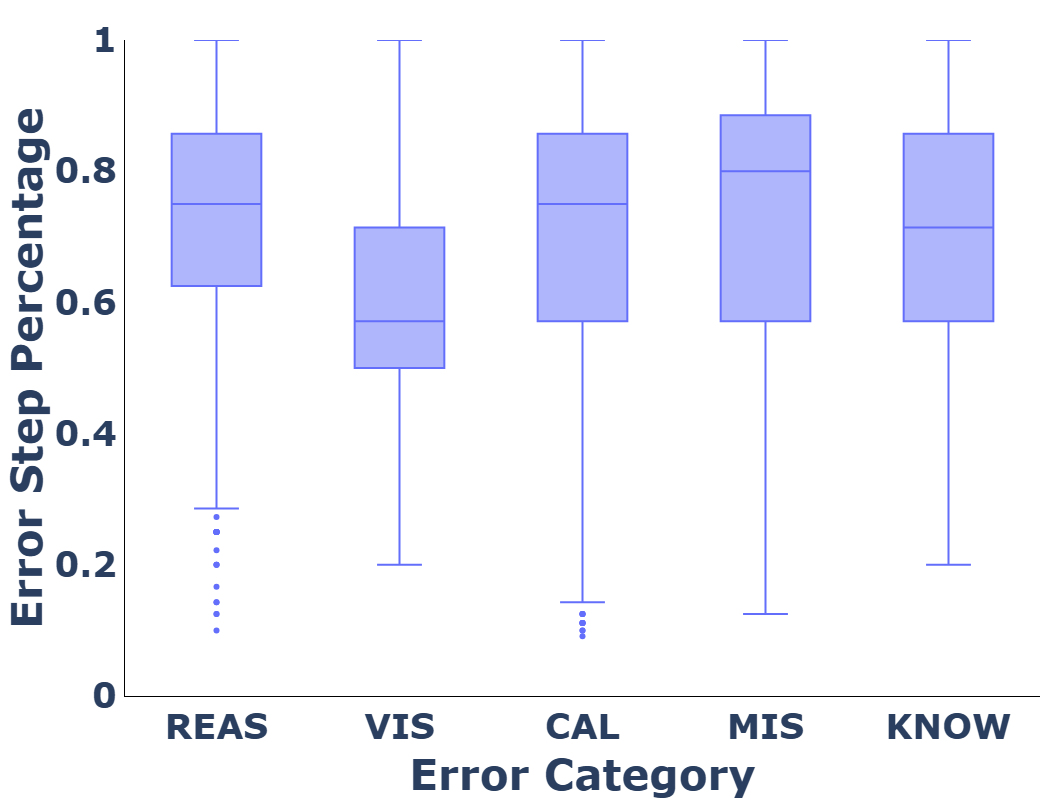}
      \caption{The error step distribution (in percentage) of error categories predicted by GPT-4o, the closed-source MLLM with the best overall performance on error detection.}
      \label{fig:step percentage gpt4o}
    \end{minipage}
\end{figure}

\section{Clarification of LLM Usage}
\label{app:llm_usage}
In the spirit of transparency, we clarify that Gemini-Pro-2.5 was utilized in the preparation of this manuscript. Its use was strictly limited to language polishing, including grammar correction, syntax refinement, and improving the overall fluency of the text. The LLM did not contribute to any of the core scientific aspects of this work. The conceptualization of the \dataset benchmark, the experimental design, the data analysis, and the interpretation of the results are entirely the original work of the human authors, who retain full responsibility for the intellectual content and integrity of this paper.




\end{document}